\documentclass[letterpaper,10pt,conference]{ieeeconf}
\IEEEoverridecommandlockouts
\usepackage{xcolor}
\usepackage{overpic}
\usepackage{amsmath,amssymb}
\usepackage{graphicx}
\usepackage{graphicx}
\usepackage{graphicx}
\usepackage{graphicx}
\graphicspath{{./img/}}
\usepackage{color}
\usepackage{url}
\usepackage{soul}
\usepackage{bm}
\usepackage{multirow}
\usepackage{multicol}
\usepackage{xspace}
\usepackage{float}
\usepackage{cite}
\usepackage{graphics}
\usepackage{subcaption}
\usepackage{tabu}
\usepackage{wrapfig}
\usepackage{algpseudocode}
\usepackage{algorithm}
\usepackage{tikz}
\usepackage{hyperref}

\newcommand{\fref}[1]{Fig.~\ref{#1}}
\newcommand{\eref}[1]{Eq.~\eqref{#1}}
\newcommand{\tref}[1]{Table~\ref{#1}}
\newcommand{\aref}[1]{Algorithm~\ref{#1}}

\DeclareMathOperator*{\argmin}{arg\!\min}
\DeclareMathOperator*{\argmax}{arg\!\max}

\newcommand\ours{JIST\xspace}
\newcommand\sbl{SBL\xspace} %sampling baseline
\newcommand\obl{OBL\xspace} %optimization baseline
\newcommand\fg{\mathcal{G}}
\newcommand\blsam{\sbl}
\newcommand\blopt{\obl}

\hypersetup{
	colorlinks=true,
	linkcolor=black,
	citecolor=black,
	urlcolor=blue
}

\newif\ifNOTARXIV % always keep
\newif\ifAPP % always keep
\APPtrue % comment out to remove appendix

\begin{document}
\pagestyle{plain}
%%%%%%%%%%%%%%%%%%%%%%%%%%%%%%%%%%%%%%%%%%%%%%%%%%%%%%%%%%%%%%%%%
\title{\bf
Joint Sampling and Trajectory Optimization over Graphs\\for Online Motion Planning
\vspace{-2mm}
}

\author{
\textbf{Kalyan Vasudev Alwala and Mustafa Mukadam}\\[2mm]
Facebook AI Research
% \thanks{}
\vspace{-1mm}
}

\twocolumn[{
	\renewcommand\twocolumn[1][]{#1}
	\maketitle
	\thispagestyle{plain}
	\begin{center}
		\vspace{-5mm}
		\centering
		\includegraphics[width=\linewidth]{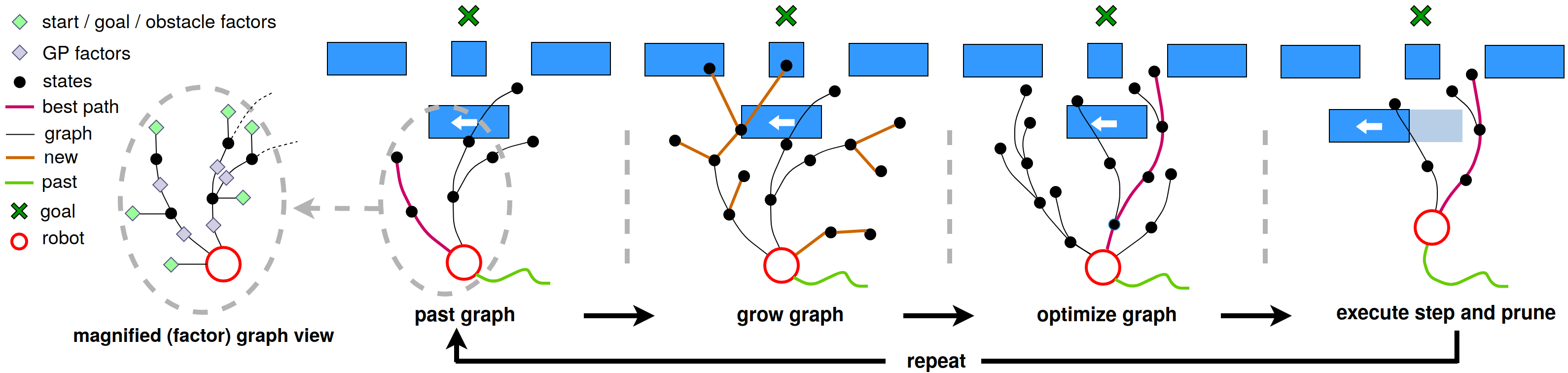}
		\captionof{figure}{\small One iteration of our approach illustrating its ability to track and switch between possible plans in an online dynamic setting.}
		\vspace{-1mm}
		\label{fig:cover}
	\end{center}
}]

%%%%%%%%%%%%%%%%%%%%%%%%%%%%%%%%%%%%%%%%%%%%%%%%%%%%%%%%%%%%%%%%%
\begin{abstract}
Among the most prevalent motion planning techniques, sampling and trajectory optimization have emerged successful due to their ability to handle tight constraints and high-dimensional systems, respectively. However, limitations in sampling in higher dimensions and local minima issues in optimization have hindered their ability to excel beyond static scenes in offline settings. Here we consider \emph{highly} dynamic environments with long horizons that necessitate a fast online solution. We present a unified approach that leverages the complementary strengths of sampling and optimization, and interleaves them both in a manner that is well suited to this challenging problem. With benchmarks in multiple synthetic and realistic simulated environments, we show that our approach performs significantly better on various metrics against baselines that employ either only sampling or only optimization.
Project page: \url{https://sites.google.com/view/jistplanner}
\end{abstract}

\IEEEpeerreviewmaketitle

%%%%%%%%%%%%%%%%%%%%%%%%%%%%%%%%%%%%%%%%%%%%%%%%%%%%%%%%%%%%%%%%%
\vspace{-2mm}
\section{Introduction and Related Work}\label{sec:intro}
\vspace{-1mm}

Many real world environments, like warehouses, hospitals, and roads, where we want our robots to operate and move about, necessitate that robots compute their motions very quickly while satisfying feasibility. We study the challenging regime of large, dense, and highly dynamic environments like these and propose a method for online motion planning in such domains.

Early research in planning with search based methods~\cite{astar,KOENIG-LPA} and their anytime variants~\cite{likhachev2004ara,likhachev2005anytime} work well on grids or graphs, the domains for which they were originally developed. These methods don't scale well to high-dimensional continuous domains and therefore have seen limited success in online or dynamic settings~\cite{ferguson2006replanning,SunYK10,mandalika2018lazy}.
Current successful and dominant strategies in motion planning can be grouped into sampling and trajectory optimization based approaches. 
Sampling methods~\cite{kavraki1996probabilistic,kuffner2000rrt,lavalle2006planning} sample the configuration space of the robot to build feasible (commonly collision free) trees or graphs that (sub)optimally connect the start and goal. Follow up work includes optimal variants~\cite{karaman2011sampling}, task relevant extensions like handling kinodynamic constraints~\cite{lavalle2001randomized,csucan2009kinodynamic}, and improving efficiency~\cite{sanchez2003single,karaman2011anytime,gammell2015batch,janson2015fast}.
They can explore the state space given sufficient time or samples and thus are well suited to low dimensional problems with tight constraints like corridors and doorways.
On the other hand, optimization methods~\cite{zucker2013chomp,toussaint2009robot,schulman2014motion,kalakrishnan2011stomp,Dong-RSS-16,Mukadam-IJRR-18} optimize an initial trajectory based on an objective function that captures some notion of optimality and feasibility. These methods are able to exploit local information to quickly find solutions and are better suited to high dimensional problems in relatively sparser settings, such as a robot arm reaching for an object on a table. Overall, both sampling and optimization notably excel at offline planning in static environments.

However their limitations become pronounced in large, dense, and dynamic settings that we consider in this work. Extending common sampling based methods to such scenarios is challenging since many samples and edges can become infeasible as the environment changes. Thus progress has been scarce for sampling methods in online~\cite{lee2017receding} and dynamic settings~\cite{ferguson2006replanning,zucker2007multipartite,aoude2013probabilistically,Otte2014RRTXRM,adiyatov2017novel}.
A potential course of action in an online setting is be to plan as quickly as possible and treat every time step as a new planning problem. This strategy is often utilized by optimization based methods and is analogous to model predictive control~\cite{tassa2008receding,allgower2012nonlinear}.
However, optimization can often get stuck in local minima as it is able to focus in the locality of only one hypothesis at a time. Multiple initializations (which may be tricky to specify) can be tried, but by incurring the penalty of added computation, linear in the number of initializations. Thus past work in employing optimization for online and dynamic problems~\cite{park2012itomp,Vannoy-RAMP,park2013real,oleynikova2016continuous} have been limited to much sparser environments.

In parallel, early planning work has also researched reactive methods~\cite{khatib1986real,quinlan1994real,chuang1998analytically,mabrouk2008solving,van2008reciprocal} that from a trajectory optimization perspective solve for a trajectory of one time step length. While this makes them extremely fast in practice, it also further exacerbates their local minima problem. Other hybrid methods, for example alternating between solving a sampling and an optimization problem~\cite{kuntz2016interleaving}, a selector that picks an initialization (straight line, random, or sample based) and an optimizer~\cite{willey2018combining}, and using optimization in generating the edges in a sampling method~\cite{choudhury2016rabit}, tend to carry over the limitations of their underlying approaches and have continued the trend in tackling only static offline settings.

A constantly evolving solution landscape makes the problem of online planning in highly dynamic environments challenging.
Regions that were previously feasible can quickly become infeasible and vice versa. In order to succeed in such settings, a planner must ideally (i) be able to continuously keep track of multiple possible hypothesis and switch between them as desired, (ii) have the ability to grow its hypothesis space in search of new possible solutions while effectively pruning less promising options, and (iii) be able to do all of this in a limited computational budget.

Our main contribution is an approach, JoInt Sampling and Trajectory optimization over graphs (\ours), that combines the complimentary strengths of exploration from sampling methods and of exploitation from optimization methods. Specifically, we employ factor graphs as the backbone data structure and extend prior work~\cite{Dong-RSS-16} (that uses chain-like graphs) to instead build more complex tree-like graphs via sampling before optimization and then pruning the graph after executing a step from the optimized solution. Repeating this process online as illustrated in \fref{fig:cover}, allows our method to achieves the three desirable properties discussed above i.e. track and switch between multiple possible plans efficiently. In various large, dense, and dynamic simulation environments with planar robots and manipulators, we show that our approach significantly outperforms a pure sampling or optimization strategy, across several metrics. For realism, in our experiments we subject the robots to limited visibility, noisy state measurements, and stochastic actions.

\begin{figure}[t]
	\centering
	\begin{subfigure}[b]{0.45\linewidth}
		\centering
		\includegraphics[width=0.99\linewidth]{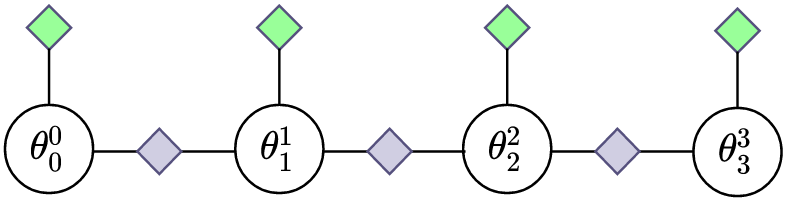}
		\vspace{0.55cm}
		\caption{\small GPMP2~\cite{Mukadam-IJRR-18}}
	\end{subfigure}
	\hspace{0.1cm}
	\begin{subfigure}[b]{0.45\linewidth}
		\centering
		\includegraphics[width=0.99\linewidth]{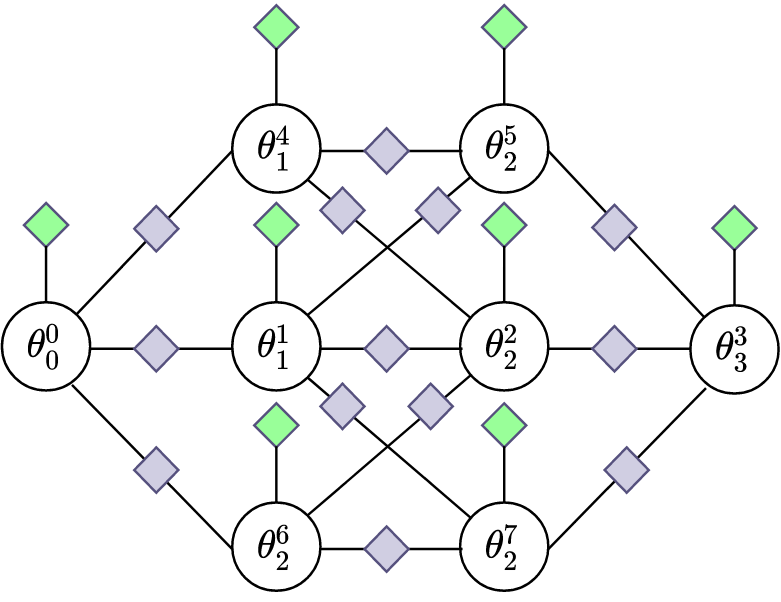}
		\caption{\small GPMP-GRAPH~\cite{Huang-ICRA-17}}
	\end{subfigure}
	\caption{\small Example factor graphs. Random variables are white circles and factors are colored shapes.}
	\label{fig:fg}
	\vspace{-1mm}
\end{figure}

%%%%%%%%%%%%%%%%%%%%%%%%%%%%%%%%%%%%%%%%%%%%%%%%%%%%%%%%%%%%%%%%%
\vspace{-2mm}
\section{The Factor Graph Backbone}\label{sec:bg}
\vspace{-1mm}

We begin by briefly reviewing the GPMP2 motion planner~\cite{Mukadam-IJRR-18} and its factor graph~\cite{kschischang2001factor} based trajectory optimization framework. Then we discuss what makes the underlying factor graph formulation a suitable choice to serve as the backbone for our approach.

The planning problem in GPMP2 is formulated as probabilistic inference on a factor graph and then solved via nonlinear least squares through the duality between inference and optimization~\cite{dellaert2006square}. This enables using factor graphs as a data structure to flexibly represent the problem, while solving them with more efficient linear algebra tools instead of message passing. Example factor graphs are shown in \fref{fig:fg} where any factor connected to a set of random variables (nodes) represents the cost function involving those variables. 

In general, the product of all conditional probabilities $\bm{f}_k$ specified by the factors gives the posterior distribution of all the random variables $\bm{\theta} = \{ \bm{\theta}^i \}$ in the graph ($i$ is the id). Likewise, the weighted sum of all cost functions $\bm{h}_k$ arising from the factors then specify the objective function whose minimization corresponds to finding the mode of the posterior
\begin{align}\label{eq:opt}
\bm{\theta}^* &= \argmax_{\bm{\theta}}  {\prod_k} \bm{f}_k
						 = \argmax_{\bm{\theta}}  {\prod_k} \exp \Big\{ \hspace{-1mm}-\frac{1}{2} \| \bm{h}_k \bm(\bm{\Theta}_k\bm) \|^2_{\bm{\Sigma}_k} \Big\}  \nonumber \\
					   &= \argmin_{\bm{\theta}} {\sum_k} \Big\{ \frac{1}{2} \bm{h}_k \bm(\bm{\Theta}_k\bm)^\top \bm{\Sigma}_k^{-1} \bm{h}_k \bm(\bm{\Theta}_k\bm) \Big\}.
\end{align}
Any cost $\bm{h}_k$ depends on $\bm{\Theta}_k$ a subset of variables and is weighted by the inverse covariance $\bm{\Sigma}_k^{-1}$ of the factor.
This optimization can be solved with an iterative scheme like Gauss-Newton by linearizing around the current solution and solving the following linear system 
\begin{equation}\label{eq:solve}
\big\{ {\textstyle\sum_k} \bm{J}^\top_k  \bm{\Sigma}_k^{-1} \bm{J} \big\} \delta \bm \theta = - {\textstyle\sum_k} \bm{J}^\top_k \bm{h}_k \bm(\bm{\Theta}_k\bm) 
\end{equation}
where Jacobian $\bm{J}_k = \partial \bm{h}_k / \partial \bm \theta$. Then the subsequent update can be performed with $\bm\theta \gets \bm \theta + \delta\bm\theta$.

In GPMP2 this factor graph takes the form of a chain as shown in \fref{fig:fg}(a) that symbolizes the trajectory where any random variable $\bm{\theta}^i_t$ with id $i$ is the state of the robot at some time step $t$. For instance, this state can be a vector of some $d$-dimensional configuration position and velocity, $\bm{\theta}^i_t \in \mathbb{R}^{2d}$. This graph consists of several factors: (i) any two consecutive states in time are connected by a GP (Gaussian Process) factor that specifies the motion model of the robot and encodes the optimality criteria to keep the trajectory smooth, (ii) every state except the first and last also connect to an obstacle factor that ensures that the state remains collision free, and (iii) the first and last state connect to a start and goal factor respectively to constrain the trajectory between start and goal locations. Similarly other factors can be added to the graph (to be part of the objective function) to accommodate other constraints like joint limits, collision avoidance between the states, etc. These and other factors we use in our work are detailed in the Appendix{\ifNOTARXIV~\cite{appendix}\fi}.

The optimization problem in \eref{eq:opt} points out that in relation to an arbitrary factor graph this formulation can be very general and not restricted to chain like graphs that encode only a single trajectory. This idea was leveraged in GPMP-GRAPH~\cite{Huang-ICRA-17} to evaluate an interconnected network of trajectories all at once. As illustrated in \fref{fig:fg}(b), it builds the factor graph with multiple chains connecting the start and goal. These chains are also connected to each other based on the spatial neighborhoods of their initialization. When this graph is optimized it results in a better exploration of the state space and thus helps finds multiple solutions or handles problems with few good local minima. This was later turned into an online approach by POSH~\cite{kolur2019online} where after taking a step on the desired path from GPMP-GRAPH the unreachable part of the graph is pruned, the environment update is received, and the graph is reoptimized to get a new path. This process is repeated until the end of the graph is reached and allows POSH to switch between different hypothesis paths as and when they became more desirable. However, not having the ability to grow the graph in any way hinders POSH in highly dynamic or long horizon problems. Additionally, the reliance on the GPMP-GRAPH template makes it tedious to define the graph structure and initialize it beyond planar settings.

The factor graph perspective does however offer a flexible way to specify any arbitrary network of trajectories with nodes as states and factors as various cost functions. The emerging sparsity can then be easily exploited to efficiently solve the optimization problem and find the optimal states encoding multiple possible solutions. In order to tackle online planning problems in highly dynamic environments, we will leverage this factor graph structure as the backbone for our approach as it will provide a way to efficiently track and switch between different hypothesis solutions. We discuss how to accomplish this in the next section.

%%%%%%%%%%%%%%%%%%%%%%%%%%%%%%%%%%%%%%%%%%%%%%%%%%%%%%%%%%%%%%%%%
\section{Combining Sampling and Optimization}

In order to efficiently plan in highly dynamic environments our key idea is to use a factor graph as the underlying data structure over which we employ a sampling strategy to build and grow the graph at any time step, then optimize, prune and repeat. We explain our approach, JoInt Sampling and Trajectory optimization over graphs (\ours) with the help of \aref{alg:alg} and a toy example shown in \fref{fig:cover}.

We begin at an intermediate iteration of \ours as shown in \fref{fig:cover} where the robot is at some current location trying to get to the goal $\bm\theta_{goal}$ after having traveled from the start state $\bm\theta_{start}$. It has a factor graph $\fg$ carried over from the previous iteration. During the first iteration this graph would be initialized with the start state as shown in line 13 of \aref{alg:alg}. As discussed in the previous section, this graph consists of various factors like GP, obstacle avoidance, goal reaching, etc that capture the motion planning objective and can be customized based on the specific task at hand. The factors we use in our experiments are described in the implementation details and elaborated in the Appendix{\ifNOTARXIV~\cite{appendix}\fi}.

In the first stage we perform exploration to discover new hypothesis solutions. We do this with the $GrowFactorGraph$ function in line 15 with an RRT-like random sampling strategy, which also leads to the underlying factor graph's tree like structure where its root is the current robot state. The critical difference compared to standard RRT~\cite{kuffner2000rrt} is that we do not check the samples or the subsequent edges for collision and instead rely on the factors in the graph and the optimization in the next stage to move them out of possible collision. Not only is this more efficient as it reduces redundant computation, but it also allows the samples to be moved by the optimizer leading to local refinement. Given the online nature and long horizons, another difference is that instead of growing the graph until we reach the goal, we grow the graph until the number of states in the graph, $\fg.size()$ hits a preset number, $node\_budget$. $node\_budget$ implicitly controls the computation needed by our method and can be tuned as a hyperparameter to achieve the best performance. As the graph is grown, factors are added in line 9, including the updated signed distance field (SDF) of the environment observed by the robot which is used for collision avoidance.

\begin{algorithm}[!t]
   \caption{\ours}
   \label{alg:alg}
    \begin{algorithmic}[1]
      \Function{$GrowFactorGraph$}{}
        \State $SDF \leftarrow env.Observe()$ 
        \While{$\fg.size() <$ $node\_budget$}
            \State $\bm\theta_{rand} \leftarrow RandomSample()$
            \State $\bm\theta_{near} \leftarrow \fg.NearestNeighbour(\bm\theta_{rand})$
            \State $\bm\theta_{new} \leftarrow ExtendState(\bm\theta_{near}, \bm\theta_{rand})$
            \State $vertex \leftarrow \fg.addVertex(\bm\theta_{near})$
            \State $edge \leftarrow \fg.addEdge(\bm\theta_{near}, \bm\theta_{new})$
            \State $\fg.addFactors(edge, vertex, SDF)$
        \EndWhile
      \EndFunction
      \Function{$Main$}{$\bm{\theta}_{start}, \bm{\theta}_{goal}$}
        \State $\fg.Initialize(\bm{\theta}_{start})$
        \While{not at $\bm{\theta}_{goal}$ or $terminal$}
            \State $\fg.GrowFactorGraph()$
            \State $\fg.Optimize()$
            \State $\bm{\theta}_{next} \leftarrow \fg.Search()$
            \State $Robot.Execute(\bm{\theta}_{next})$
            \State $\bm{\hat\theta}_{next} \leftarrow Robot.MeasureState()$ 
            \State $\fg.PruneUnreachable(\bm{\hat\theta}_{next} )$
        \EndWhile
       \EndFunction
\end{algorithmic}
\end{algorithm}

In the next stage we exploit the local information in line 16 and optimize the new graph that consists of the old graph and the newly sampled portion. The optimization is efficiently performed with the iterative scheme in \eref{eq:opt}-\eqref{eq:solve} until convergence to get all the locally optimal states in the graph. This enables our approach to track multiple possible hypothesis paths. Now we can choose the next best step or steps to be executed by searching over the converged graph and leads to a potential switch in the current path compared to previous iteration as shown in \fref{fig:cover}. Finally, the (noisy) robot state is measured after (stochastic) execution and the unreachable parts of the graph are pruned. This procedure is repeated during every iteration of \ours until the goal is reached or another terminal condition (like collision) is met.

Thus our method \ours explores via sampling and exploits via efficient optimization and combines the benefits of both by leveraging a factor graph backbone that dynamically grows and shrinks between iterations. This allows it to explore, track, and switch between possible solutions as the environment changes and the solution space evolves. These features are desirable in planning for environment that are large, dense, and highly dynamic. \ours also offers flexibility in choosing any sampling strategy, objectives for the cost, optimization, as well as the method to search over the graph, based on what is relevant to the task and application.

\begin{figure*}[!t]
\centering
\includegraphics[trim=53 32 45 25,clip,width=0.24\linewidth]{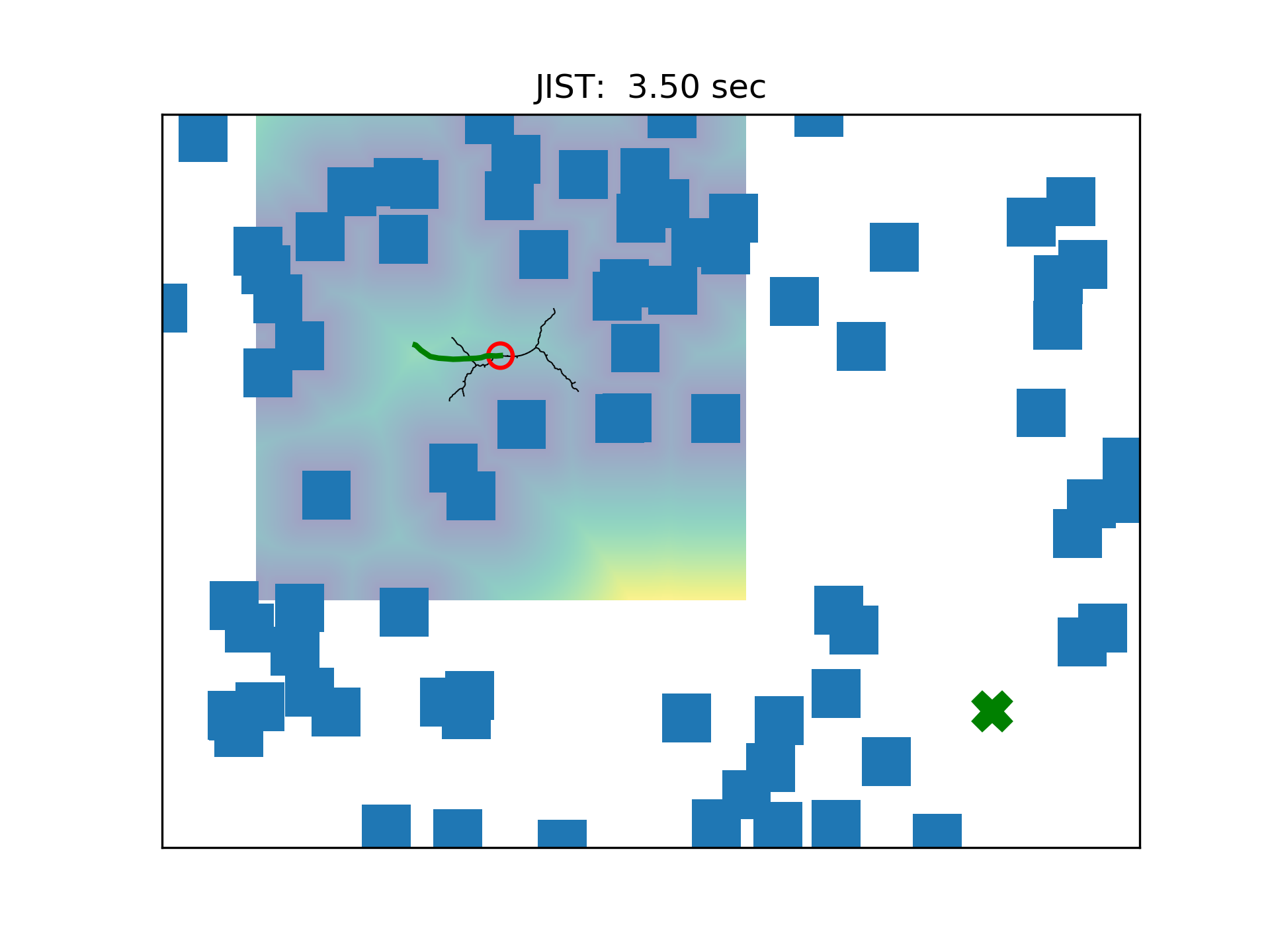}
\includegraphics[trim=53 32 45 25,clip,width=0.24\linewidth]{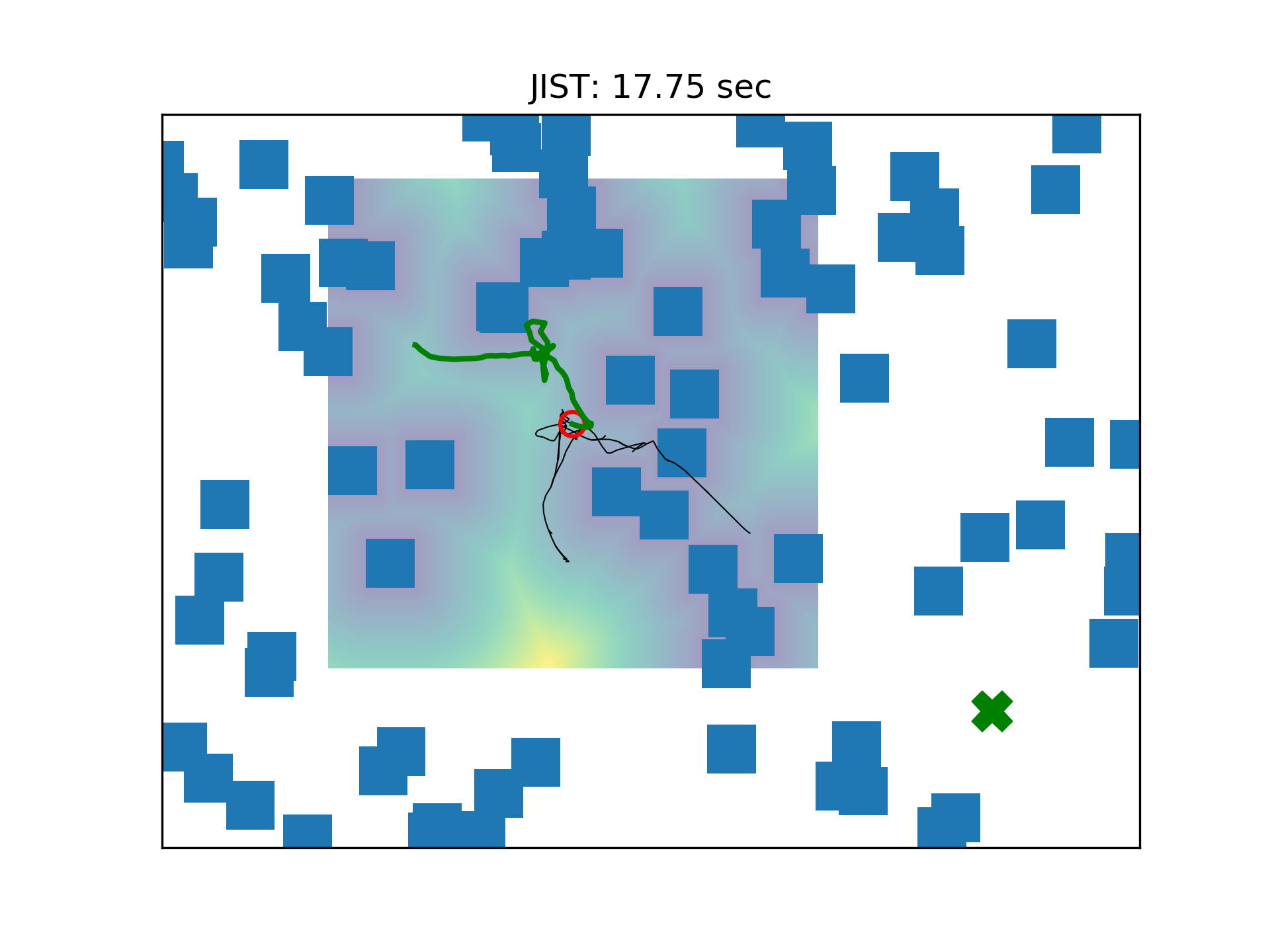}
\includegraphics[trim=53 32 45 25,clip,width=0.24\linewidth]{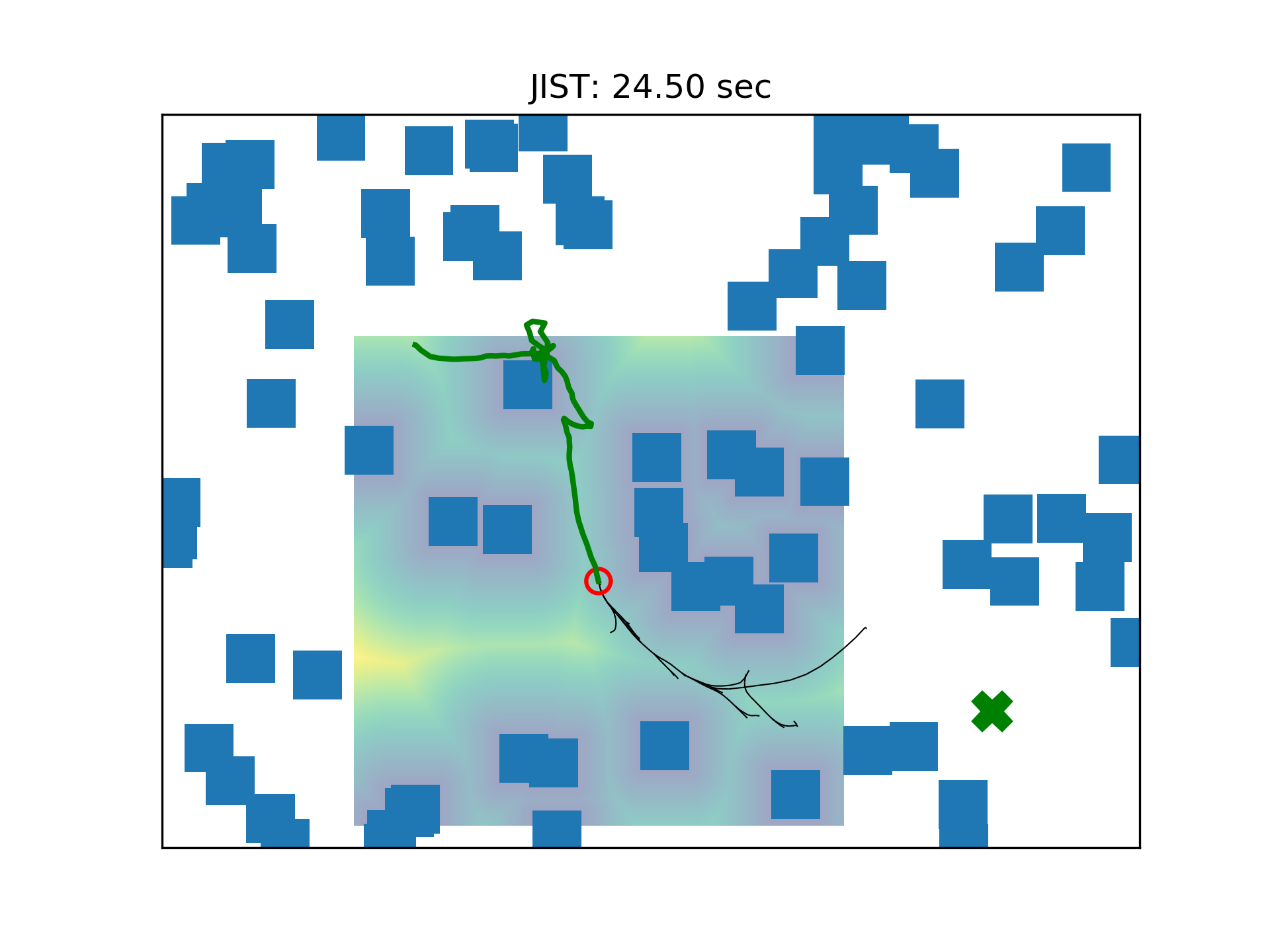}
\includegraphics[trim=53 32 45 25,clip,width=0.24\linewidth]{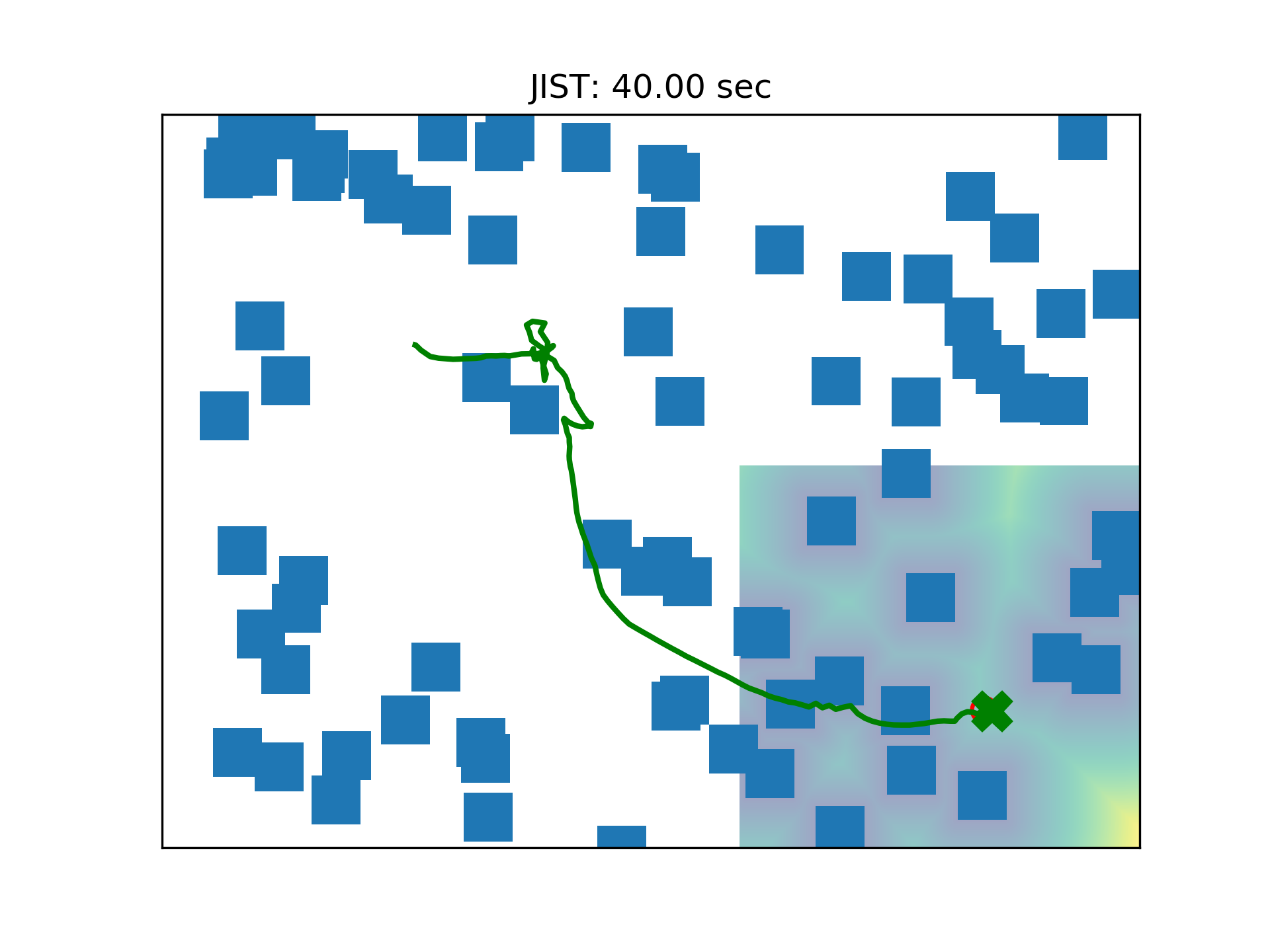}

\includegraphics[trim=53 32 45 25,clip,width=0.24\linewidth]{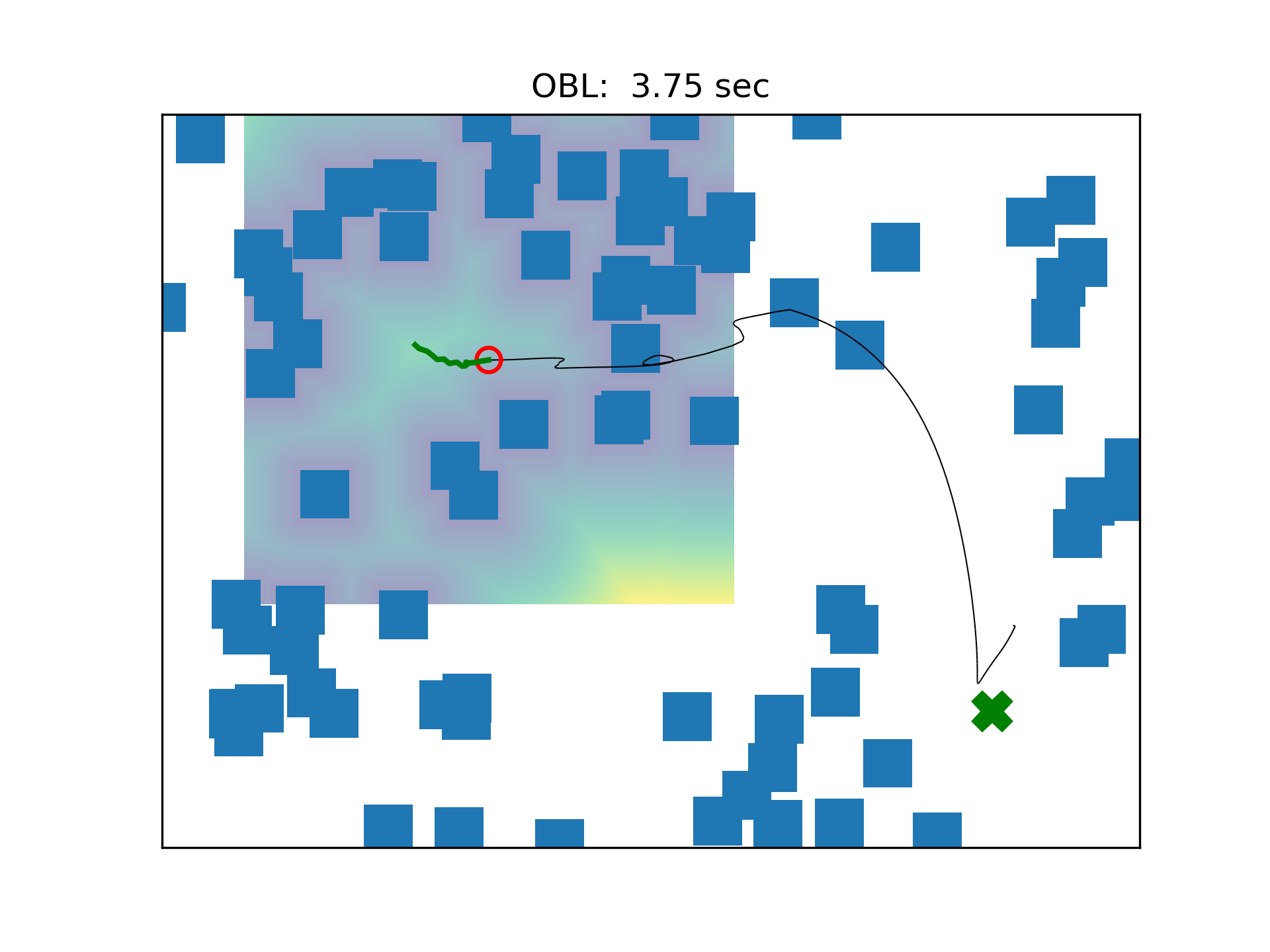}
\includegraphics[trim=53 32 45 25,clip,width=0.24\linewidth]{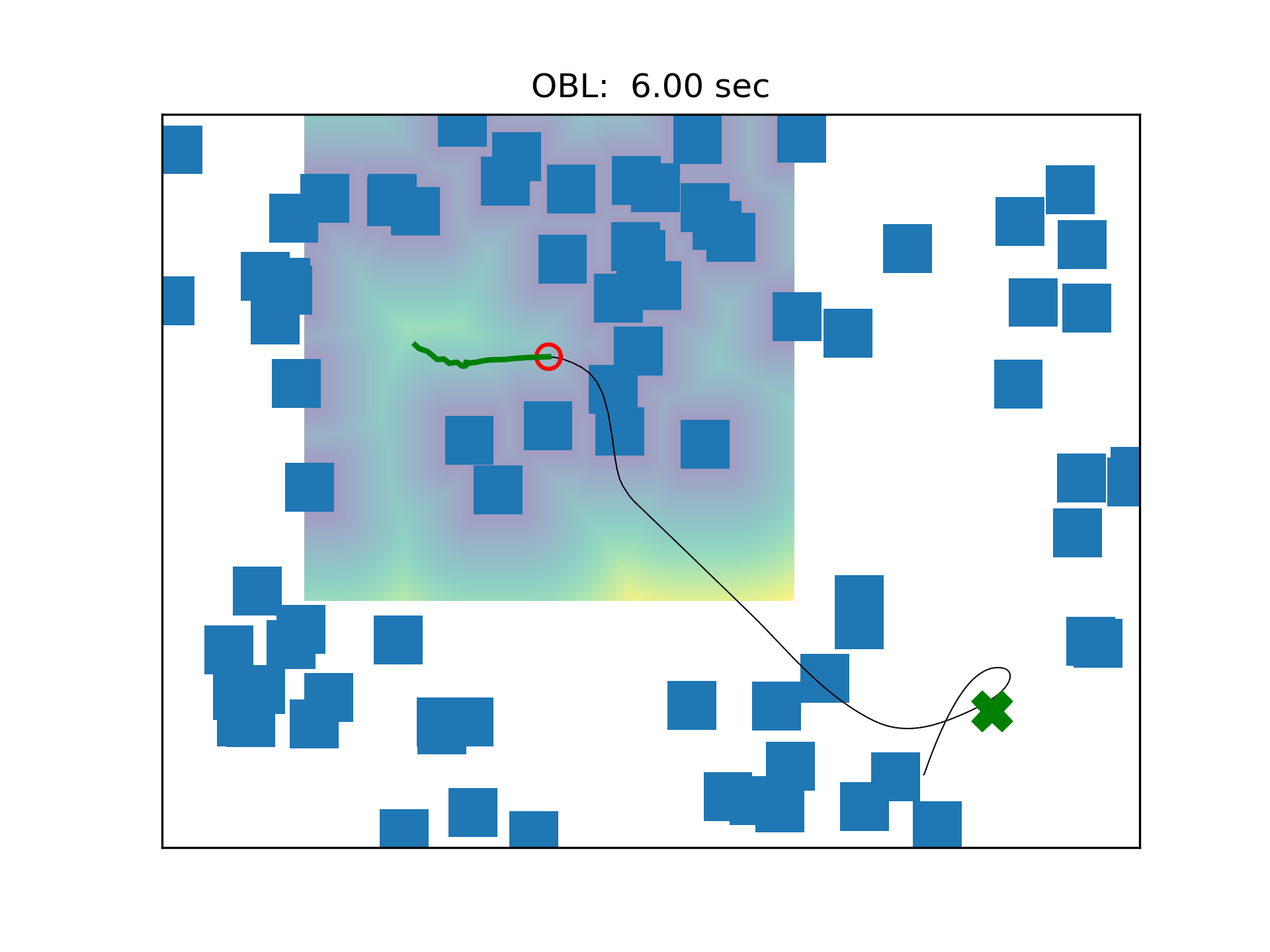}
\includegraphics[trim=53 32 45 25,clip,width=0.24\linewidth]{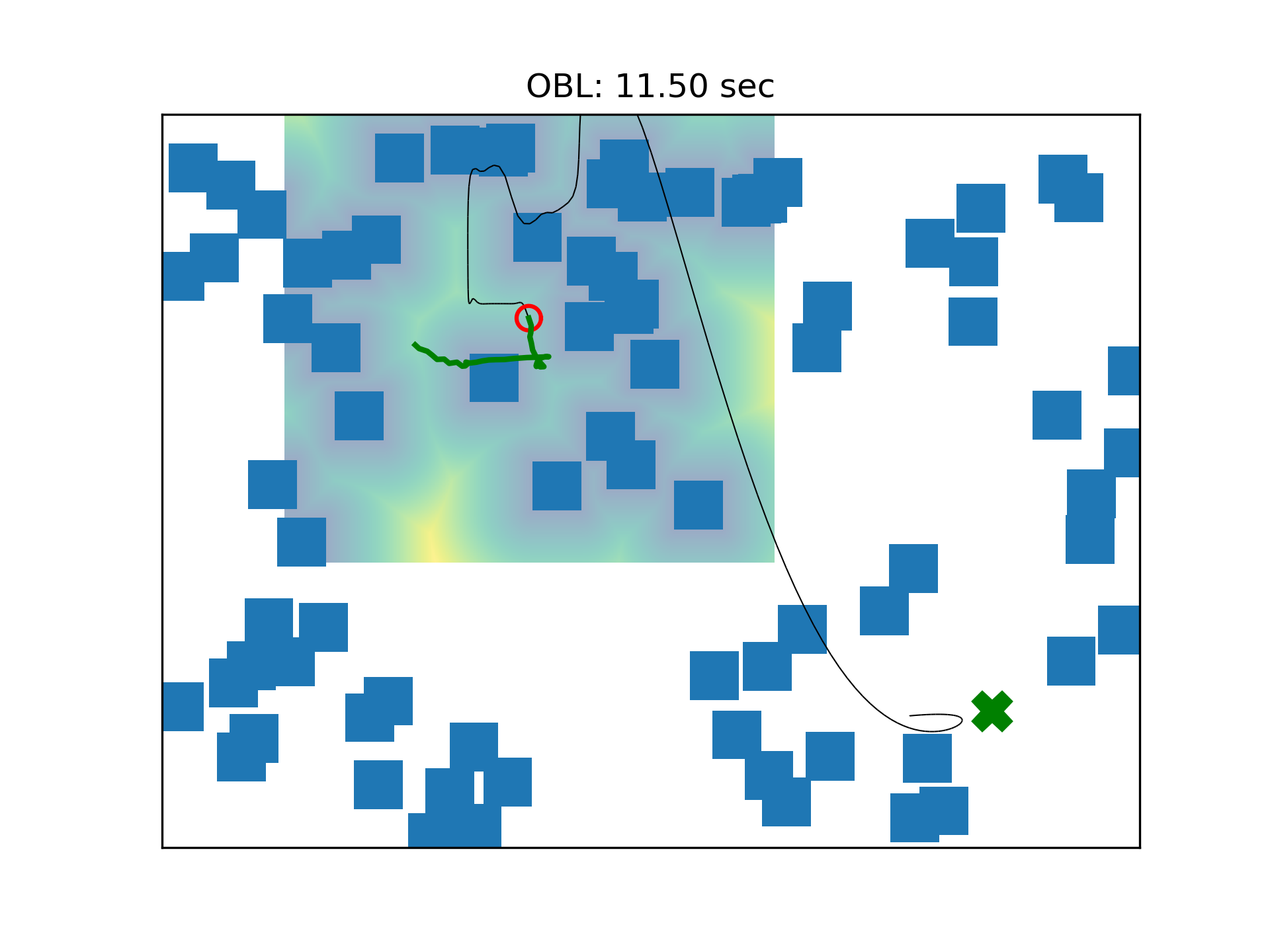}
\includegraphics[trim=53 32 45 25,clip,width=0.24\linewidth]{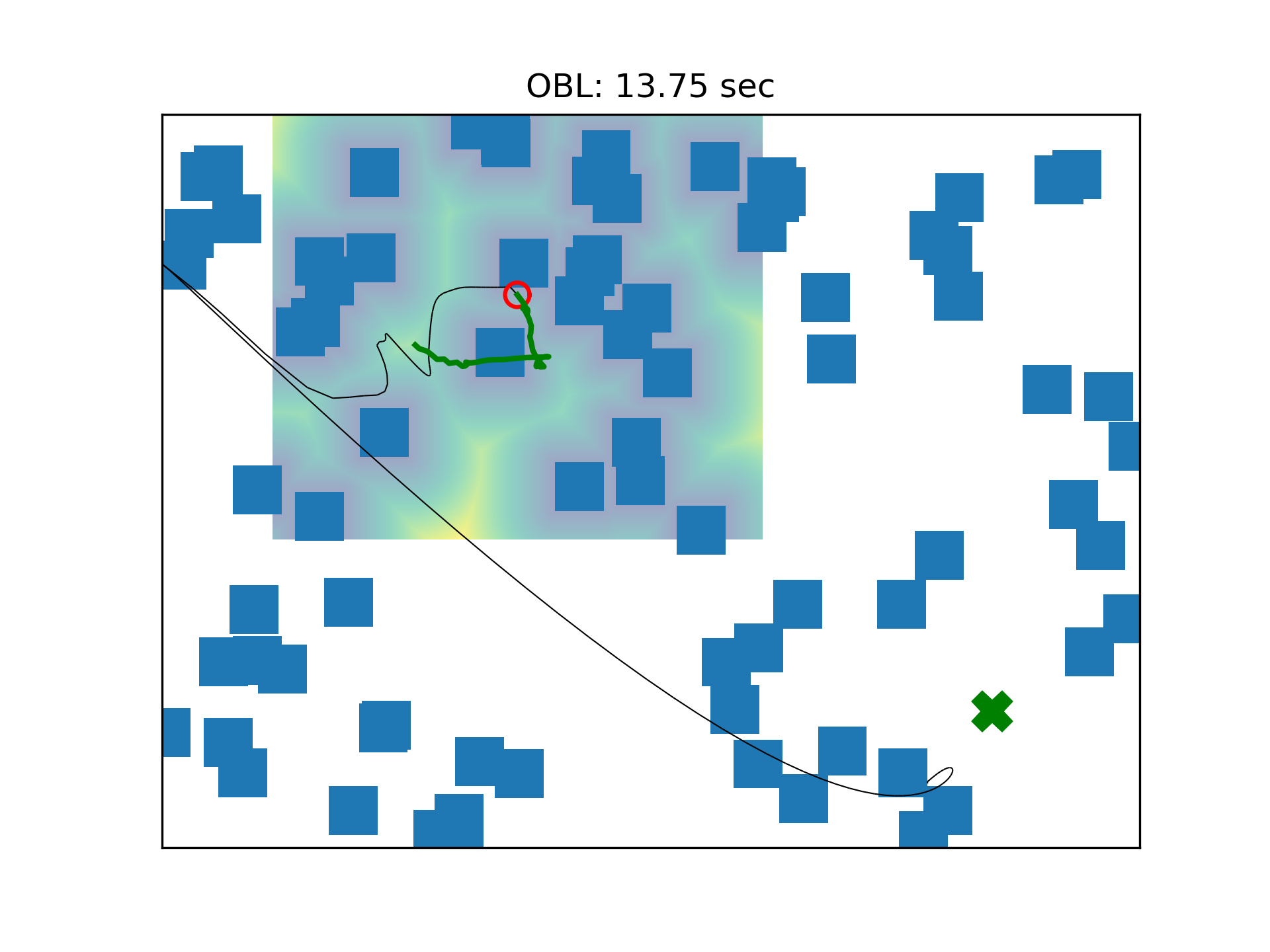}

\includegraphics[trim=53 32 45 25,clip,width=0.24\linewidth]{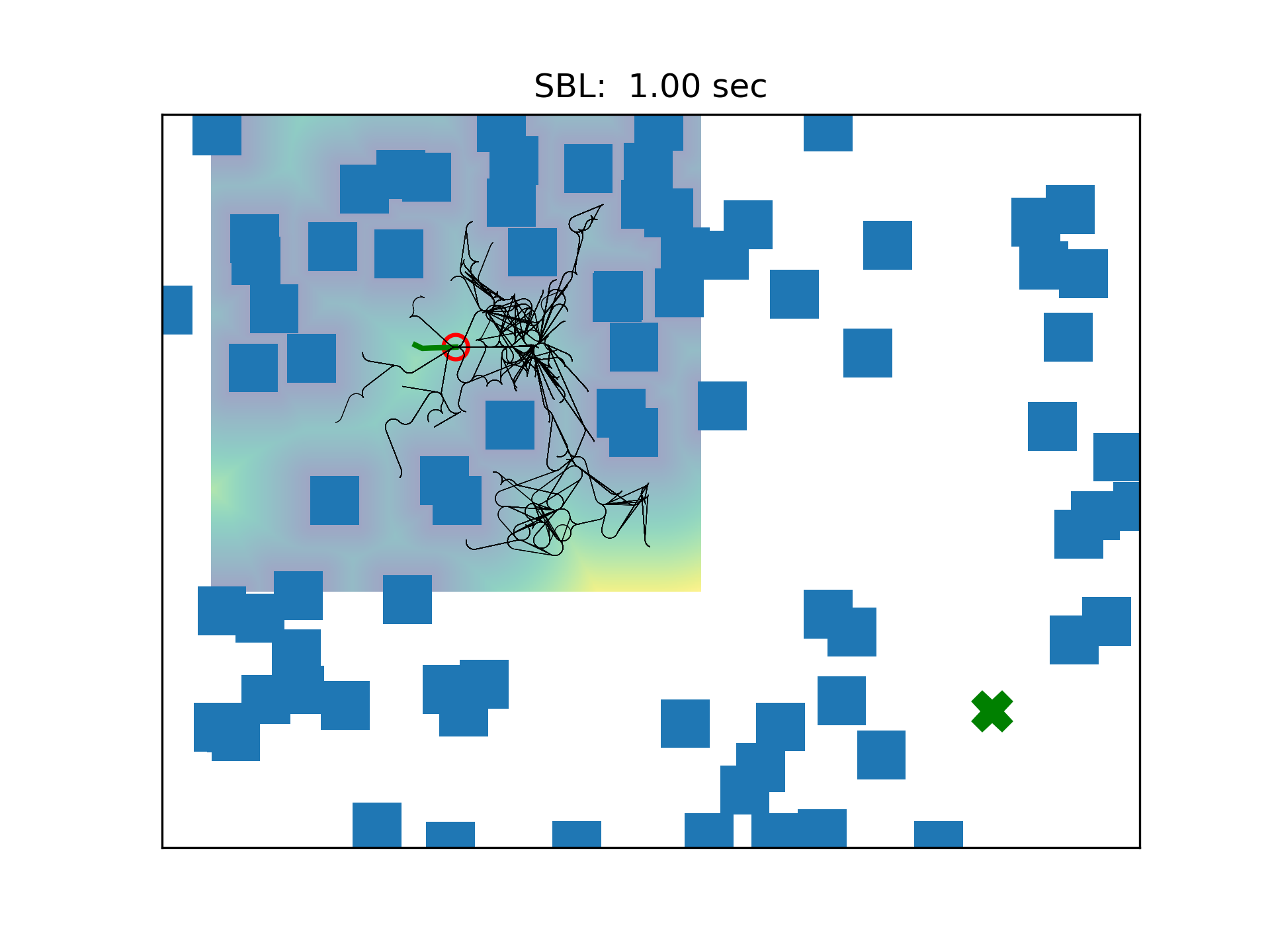}
\includegraphics[trim=53 32 45 25,clip,width=0.24\linewidth]{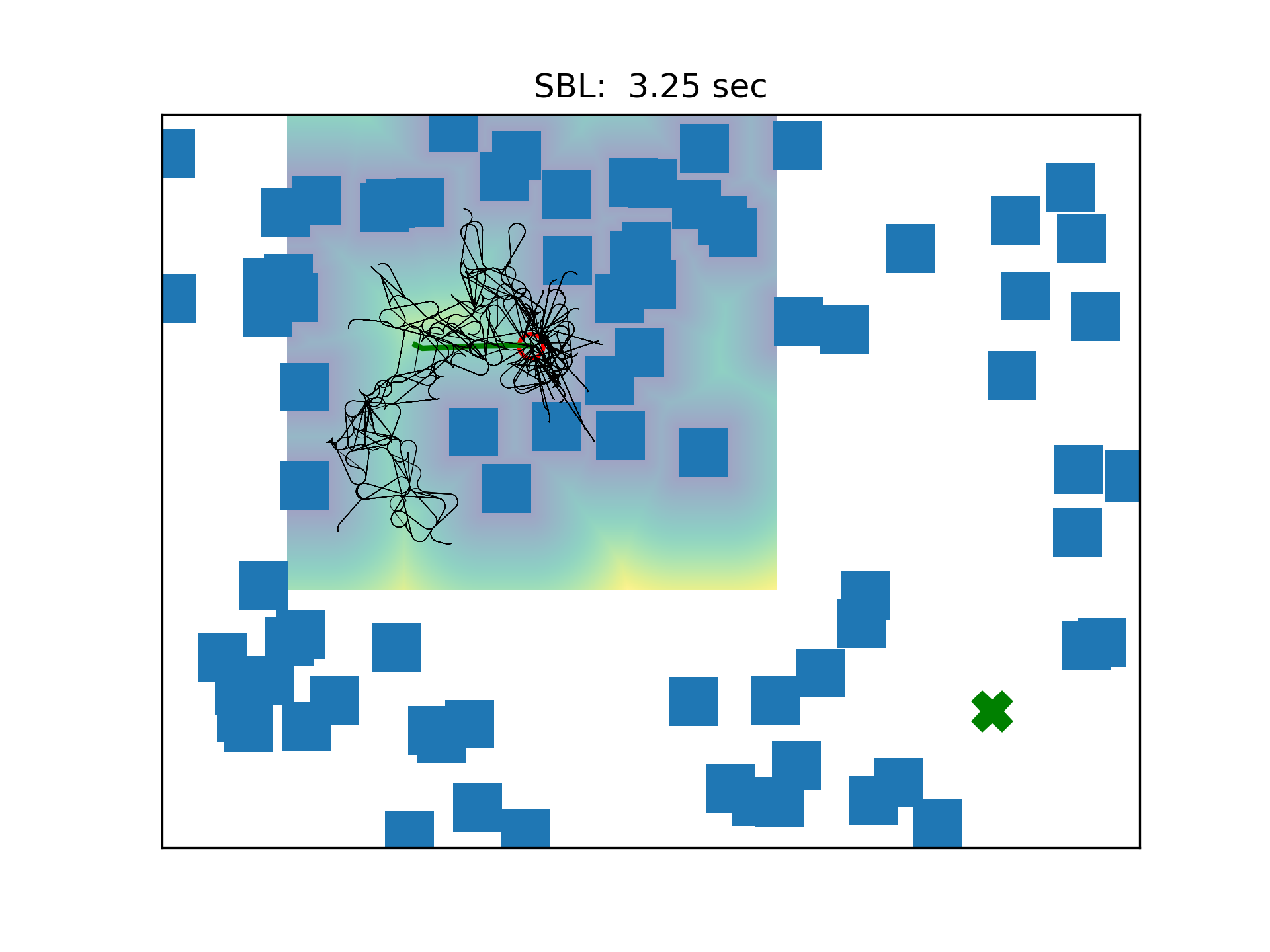}
\includegraphics[trim=53 32 45 25,clip,width=0.24\linewidth]{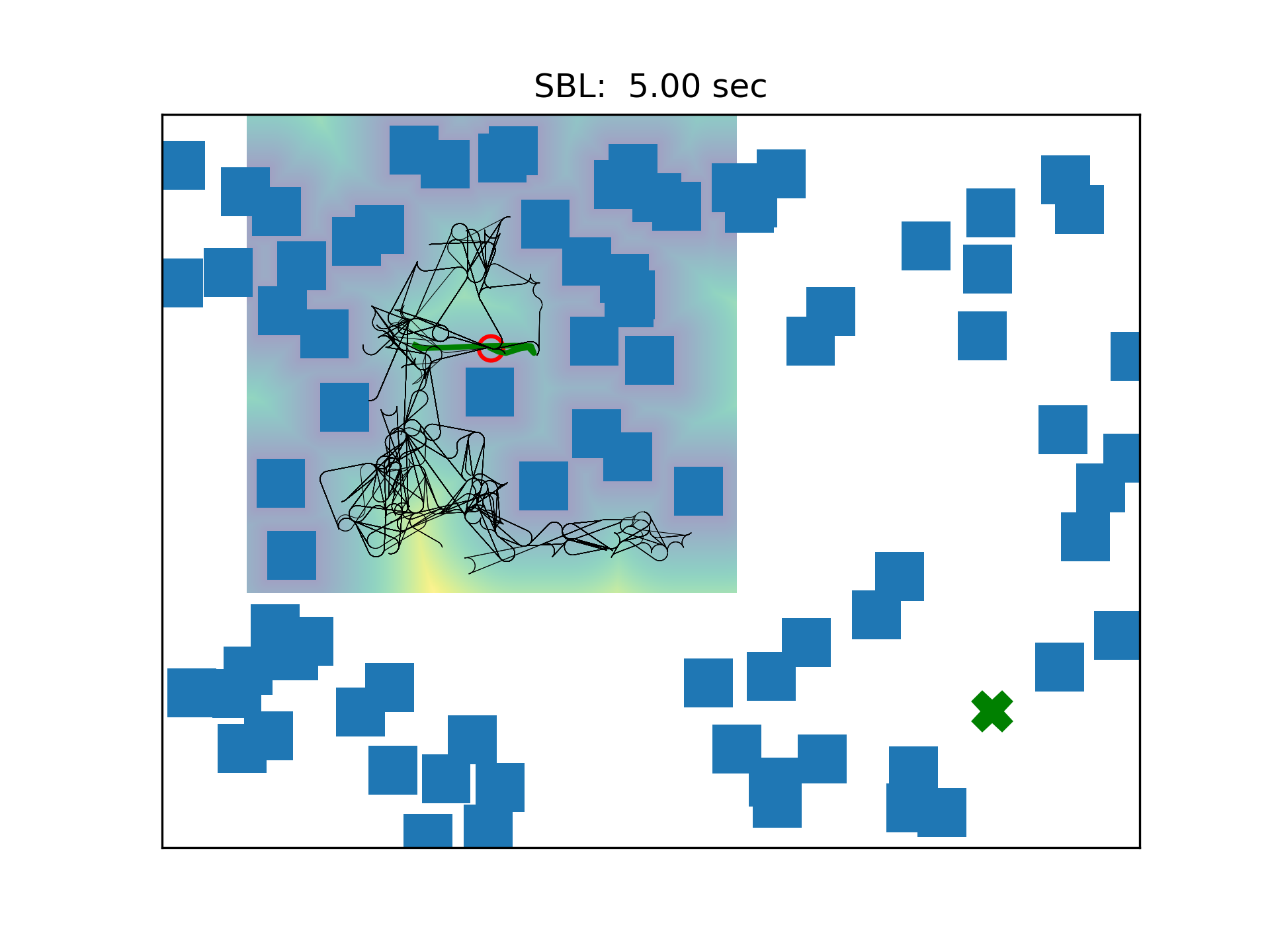}
\includegraphics[trim=53 32 45 25,clip,width=0.24\linewidth]{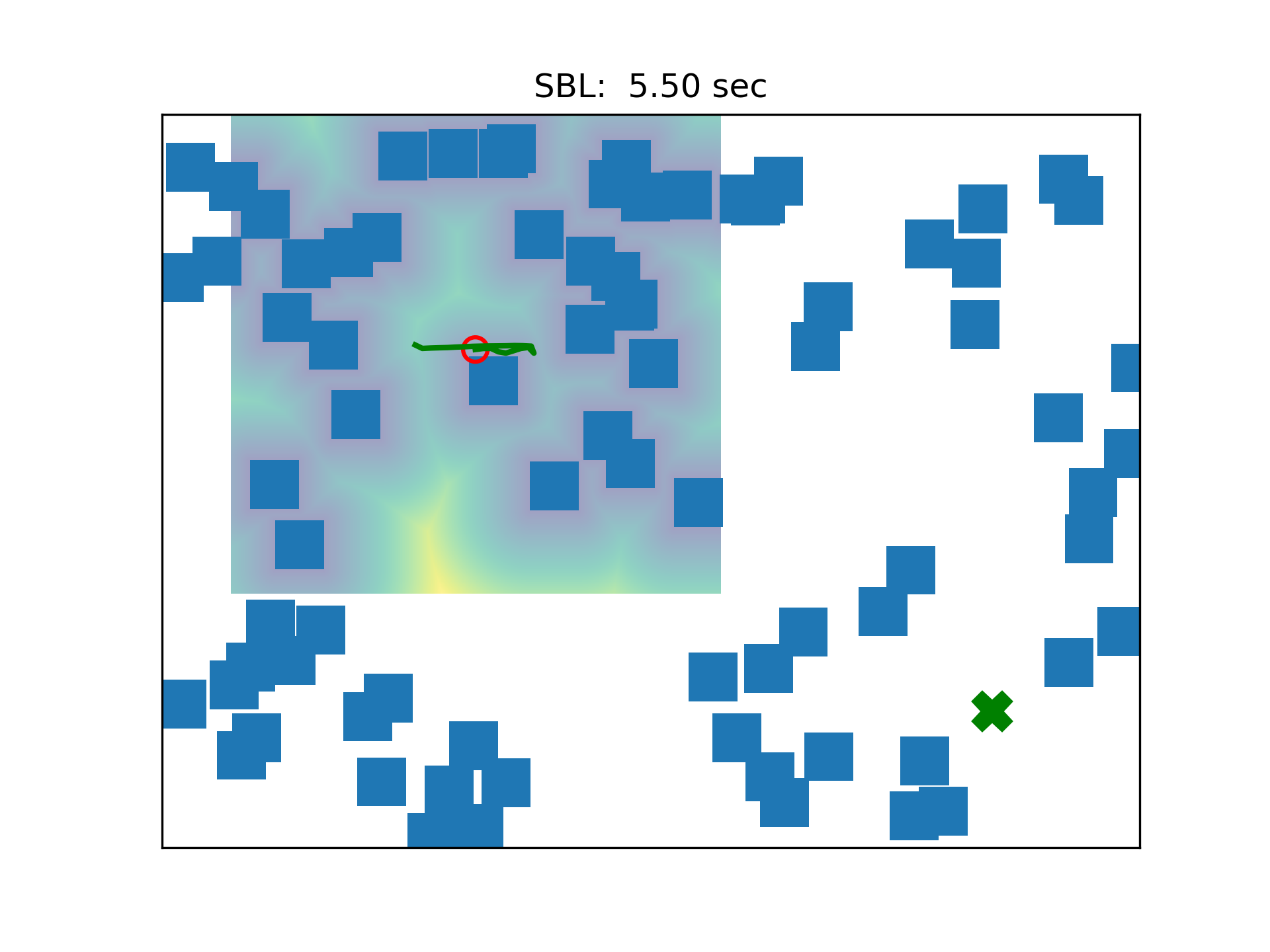}

\captionof{figure}{\small Same 2D Forest benchmark example (left to right) where \ours (ours) (top row) successfully reaches the goal while \blopt (middle row) and \blsam (bottom row) terminate in collision.
}
\label{fig:random}
\vspace{-6mm}
\end{figure*}

\subsection{Implementation Details}

We implemented \ours using the GTSAM~\cite{frank2012factor} and the GPMP2~\cite{Mukadam-IJRR-18} libraries. While sampling states and building the factor graph, we add the following factors on the states and between states that encode the motion planning objectives. We apply a Gaussian factor on the current state of the robot to encode the measurement and anchor it as the root of the tree structured graph. Consecutive states in time are connected by the Gaussian process (GP) factor that encodes smoothness of the trajectory (i.e. the motion model of the robot). For collision avoidance we use unary obstacle factors~\cite{Mukadam-IJRR-18} on all states and binary obstacle factors between consecutive states that use GP interpolation and ensure path safety~\cite{Mukadam-IJRR-18}. For goal reaching we adopt the goal factor from~\cite{Mukadam-ICRA-17} and similarly have a goal factor on every state except the current state which allows the graph to drive all paths toward the goal weighted by how far the robot currently is from the goal. We also use domain specific factors that enforce different constraints like velocity limits, joint limits, self-collision, etc and identify them in the next section where we describe each domain. All factors used in our work are detailed in the Appendix{\ifNOTARXIV~\cite{appendix}\fi}.

\begin{figure*}[!t]
  \centering
  \includegraphics[trim=177 35 165 25,clip,width=0.08 \linewidth]{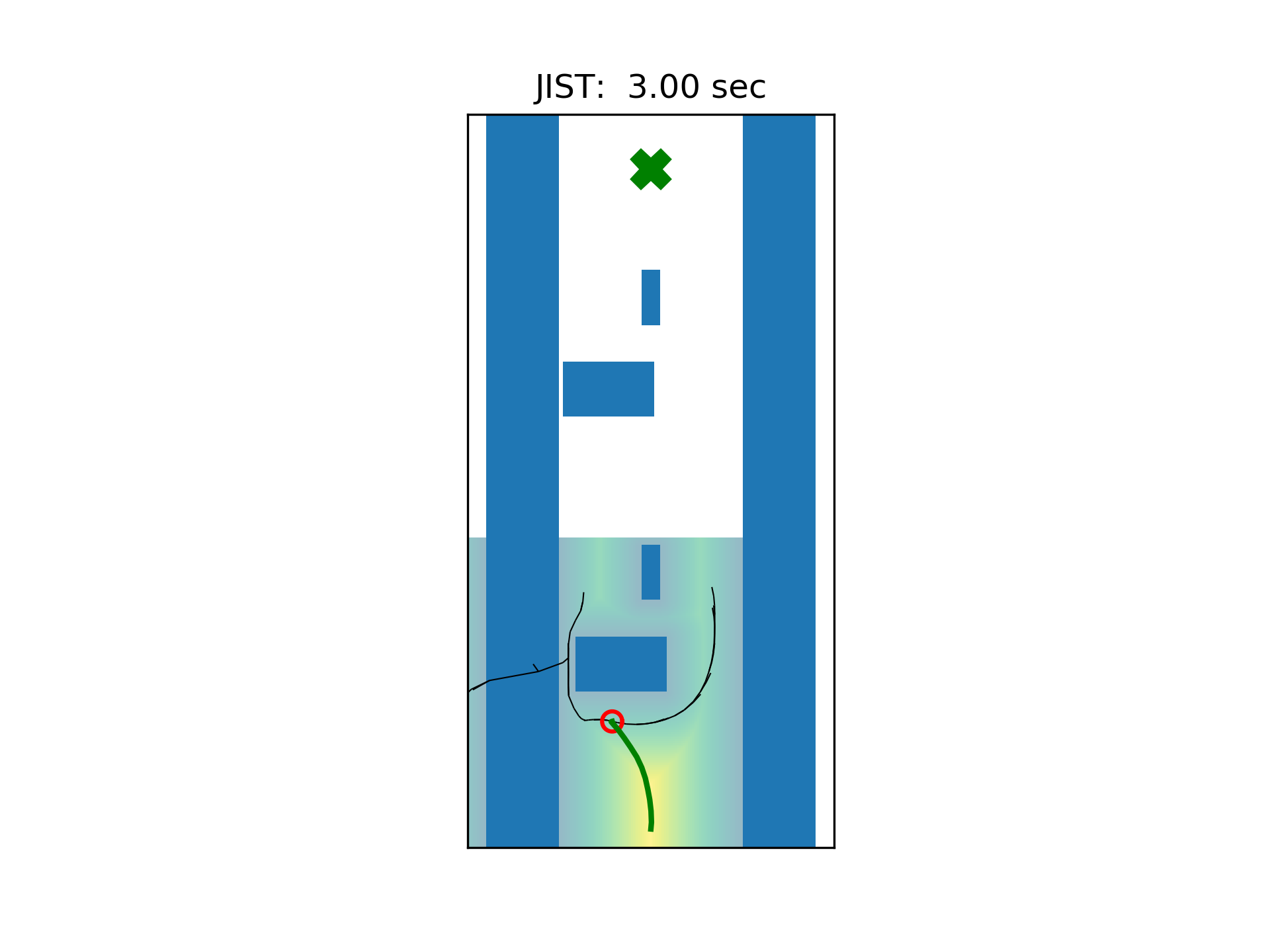}
  \hspace{-0.1cm}
  \includegraphics[trim=177 35 165 25,clip,width=0.08\linewidth]{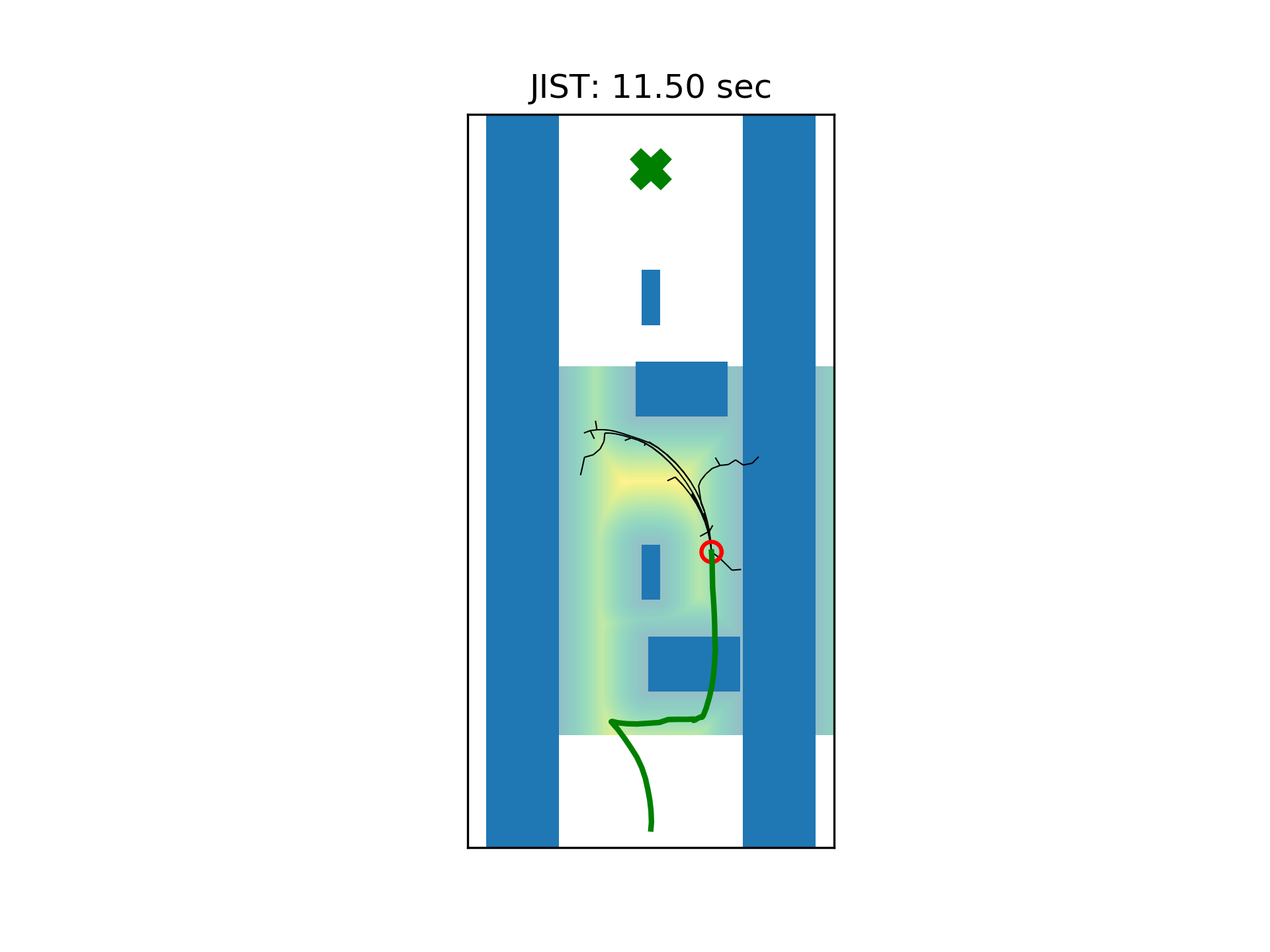}
  \hspace{-0.1cm}
  \includegraphics[trim=177 35 165 25,clip,width=0.08\linewidth]{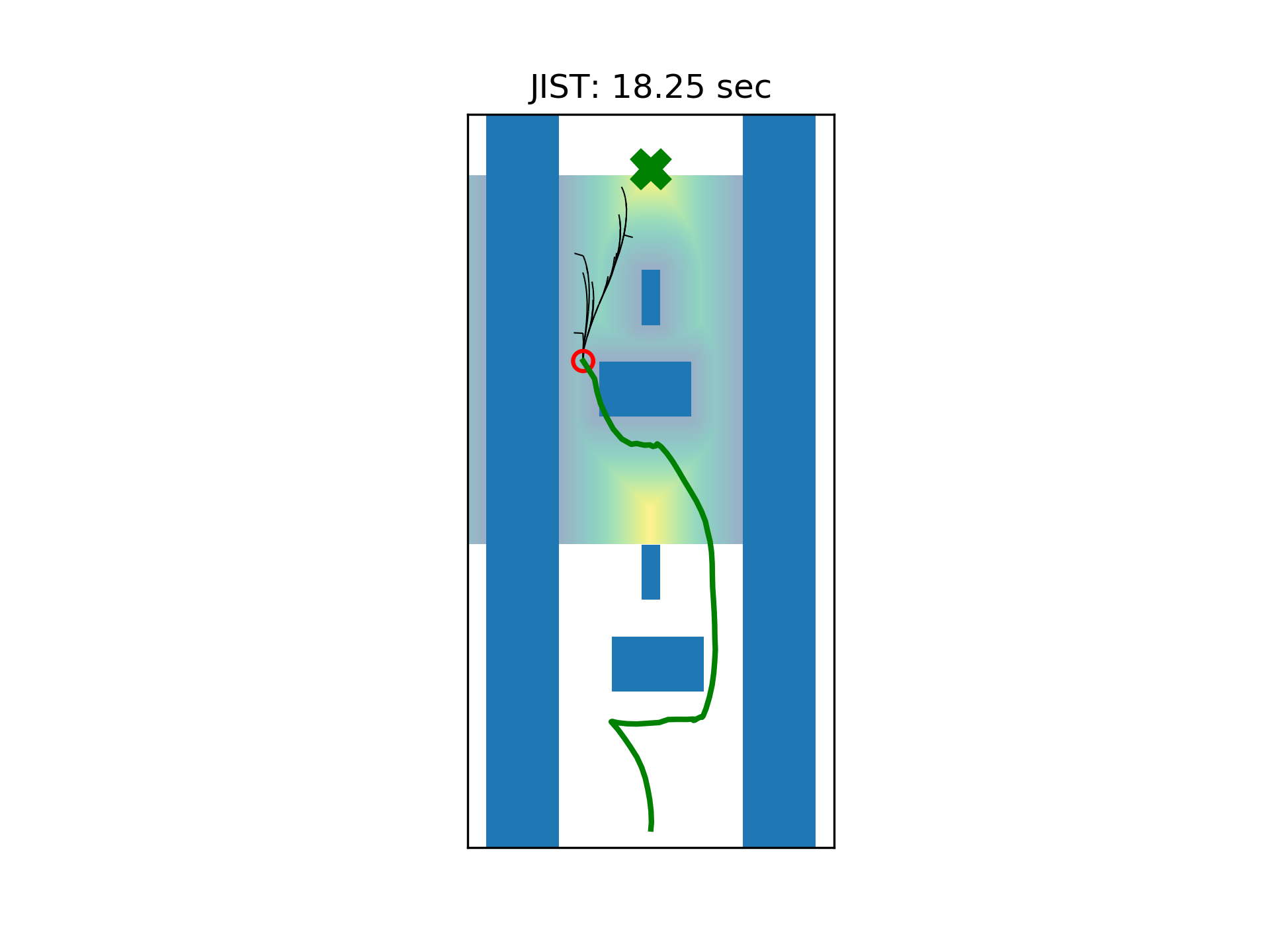}
  \hspace{-0.1cm}
  \includegraphics[trim=177 35 165 25,clip,width=0.08\linewidth]{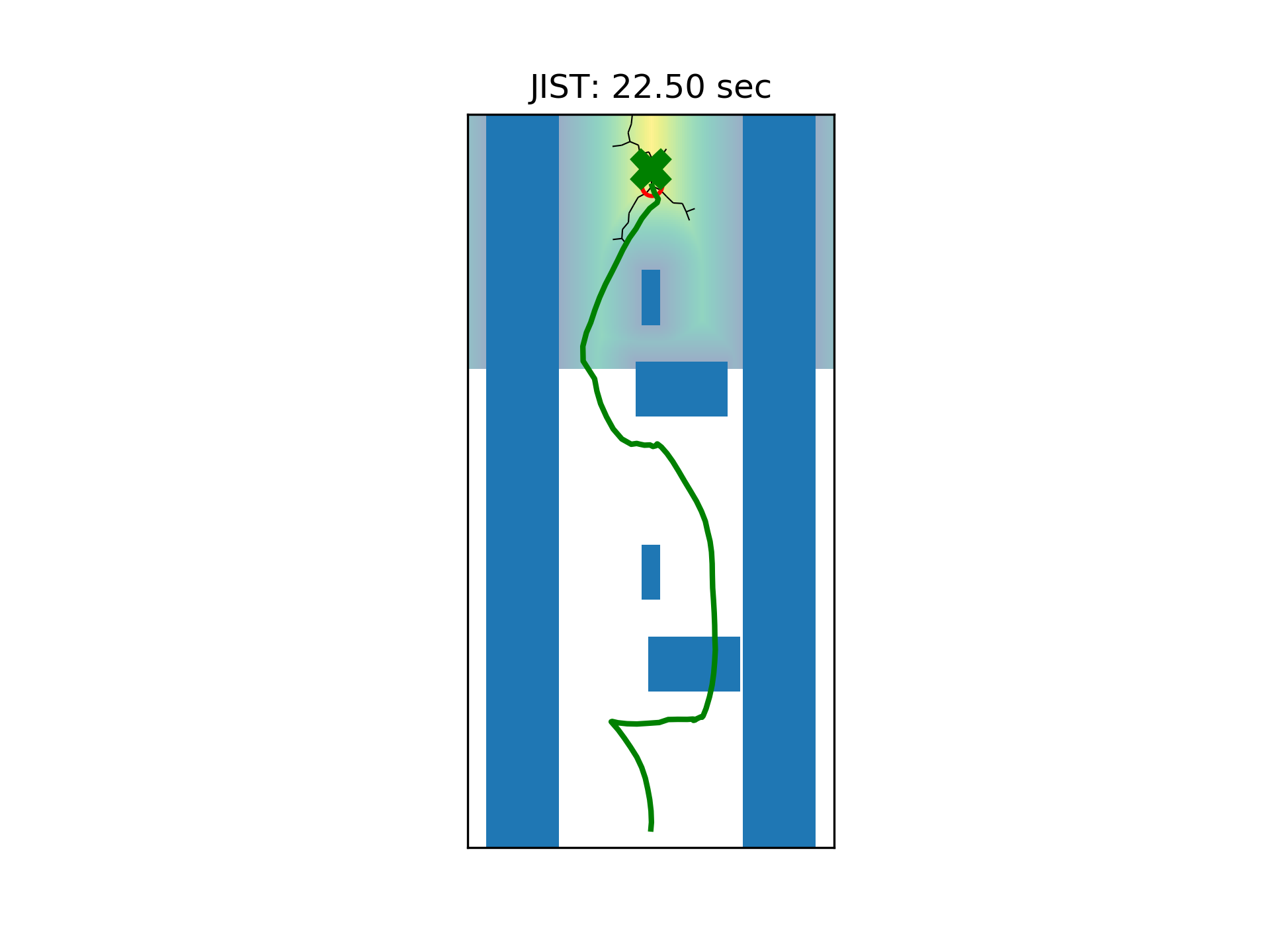}
  \hspace{0.5cm}
  \includegraphics[trim=177 35 165 25,clip,width=0.08\linewidth]{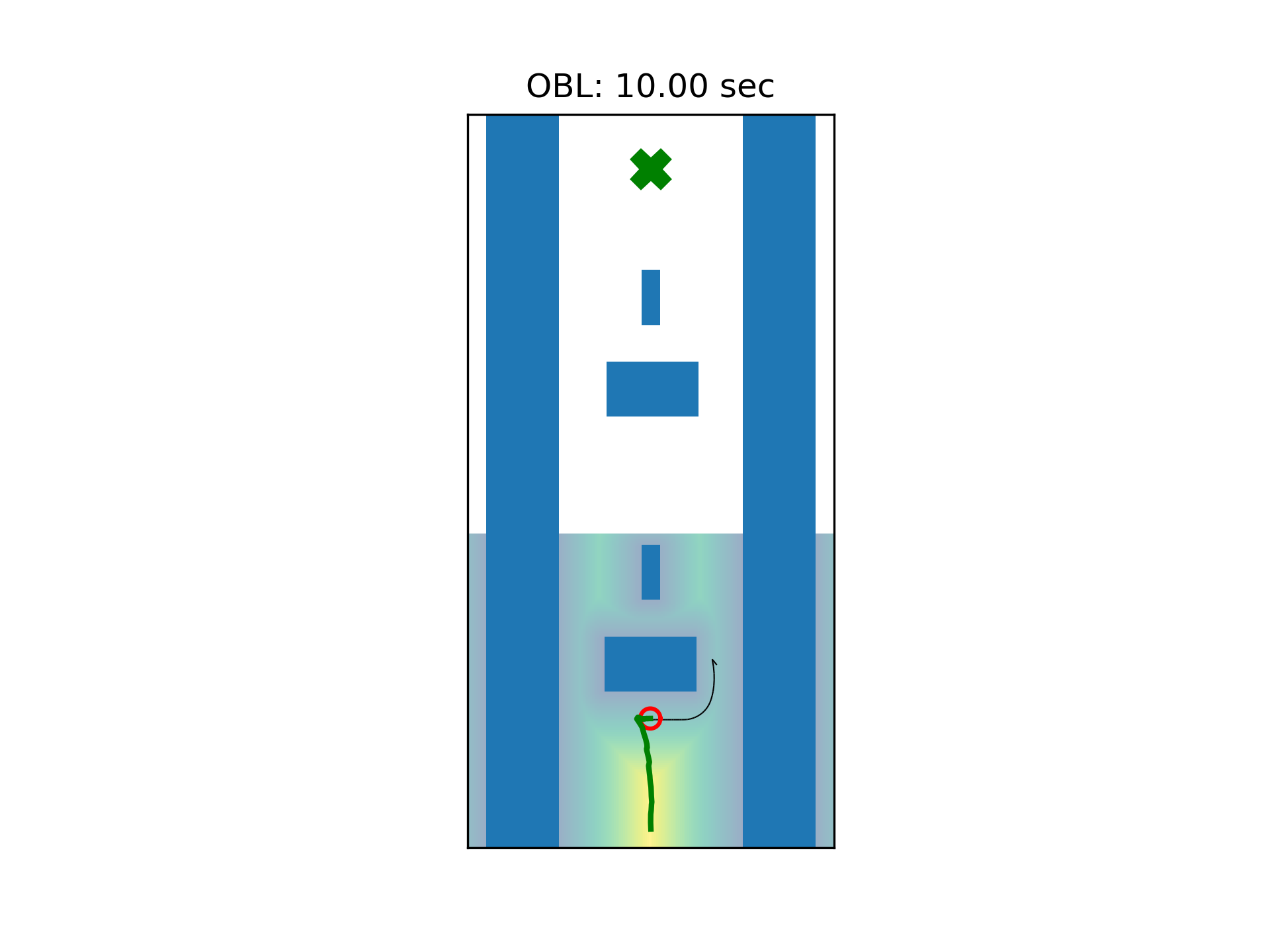}
  \hspace{-0.1cm}
  \includegraphics[trim=177 35 165 25,clip,width=0.08\linewidth]{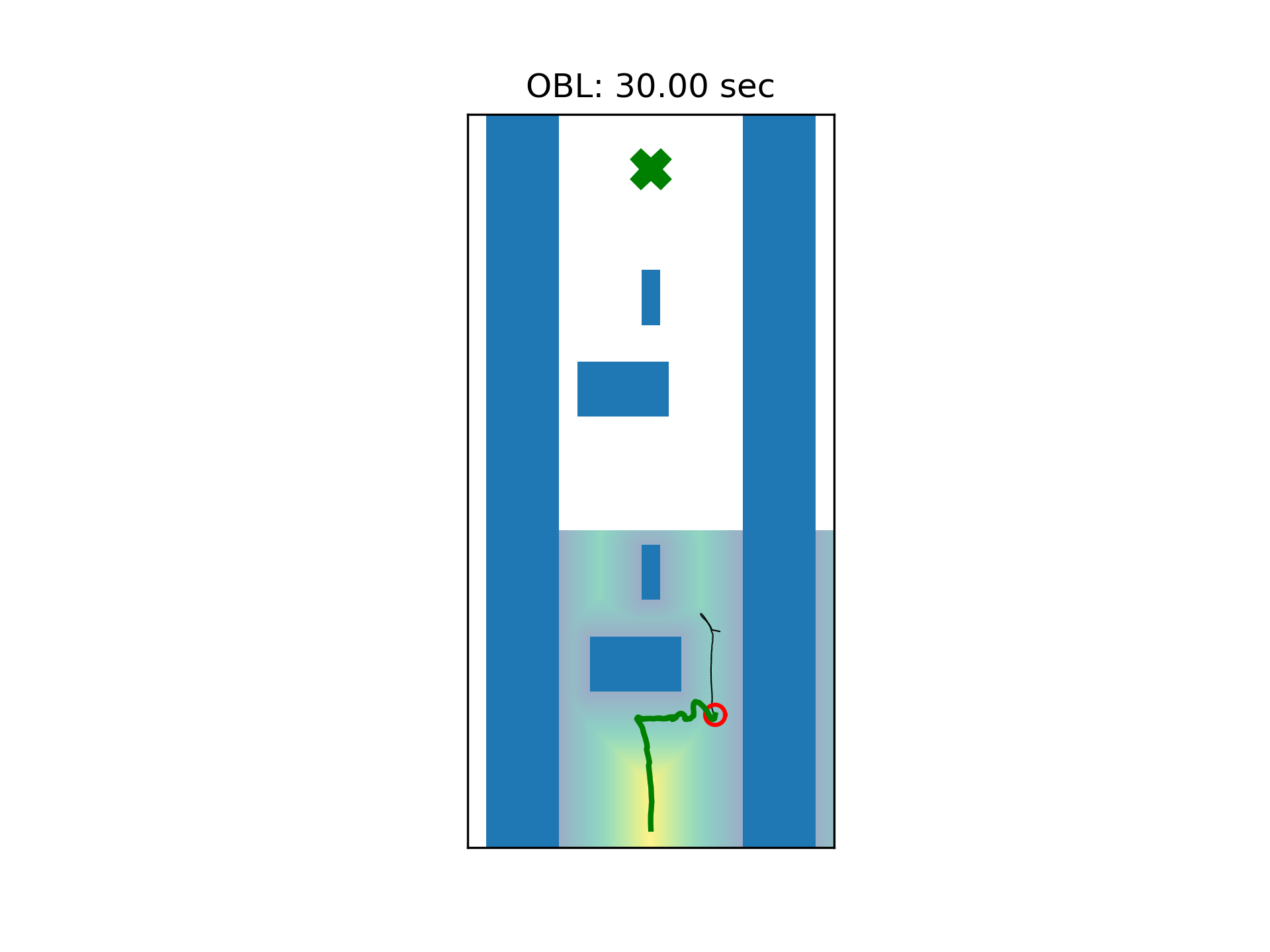}
  \hspace{-0.1cm}
  \includegraphics[trim=177 35 165 25,clip,width=0.08\linewidth]{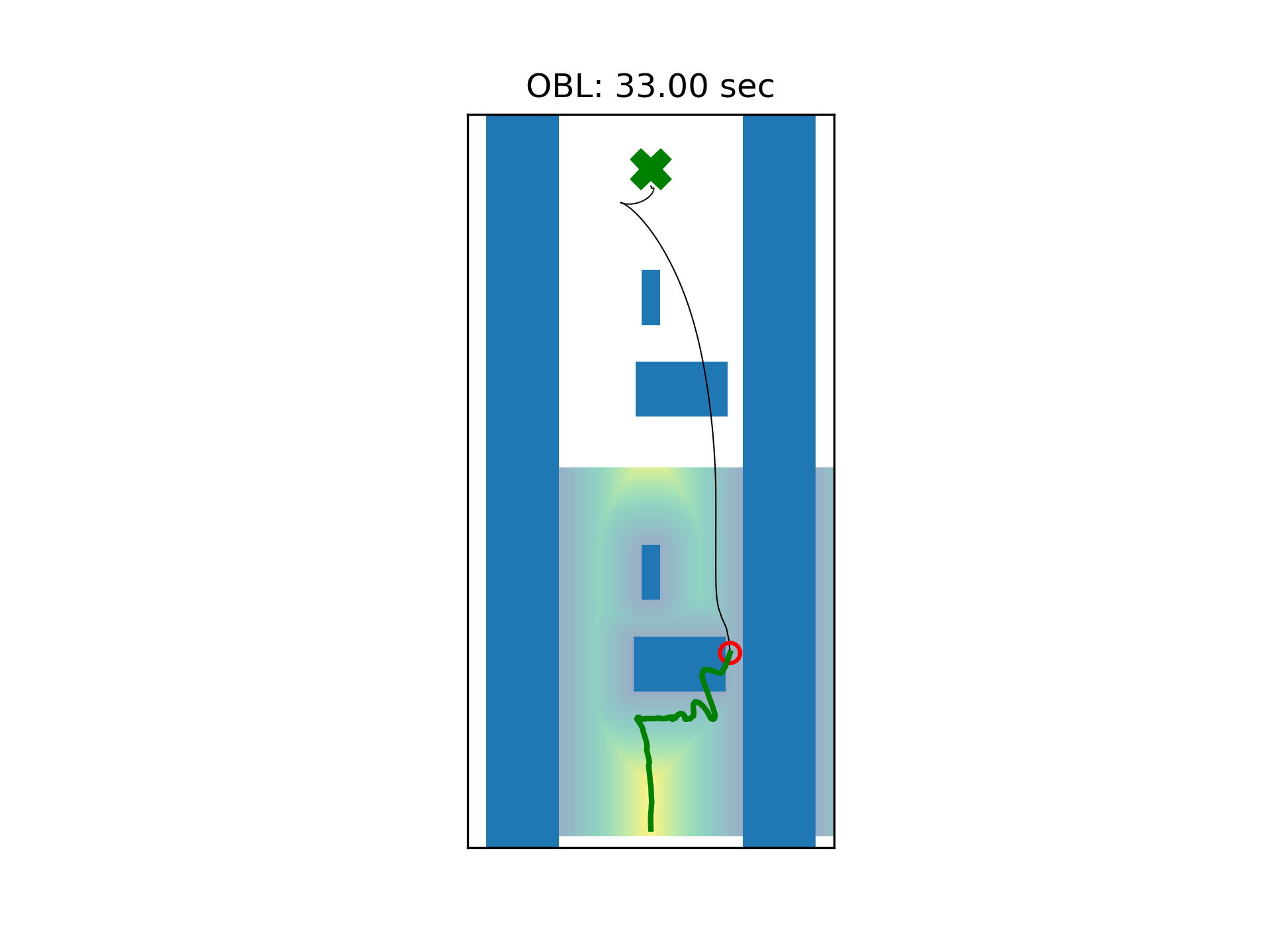}
  \hspace{0.5cm}
  \includegraphics[trim=177 35 165 25,clip,width=0.08\linewidth]{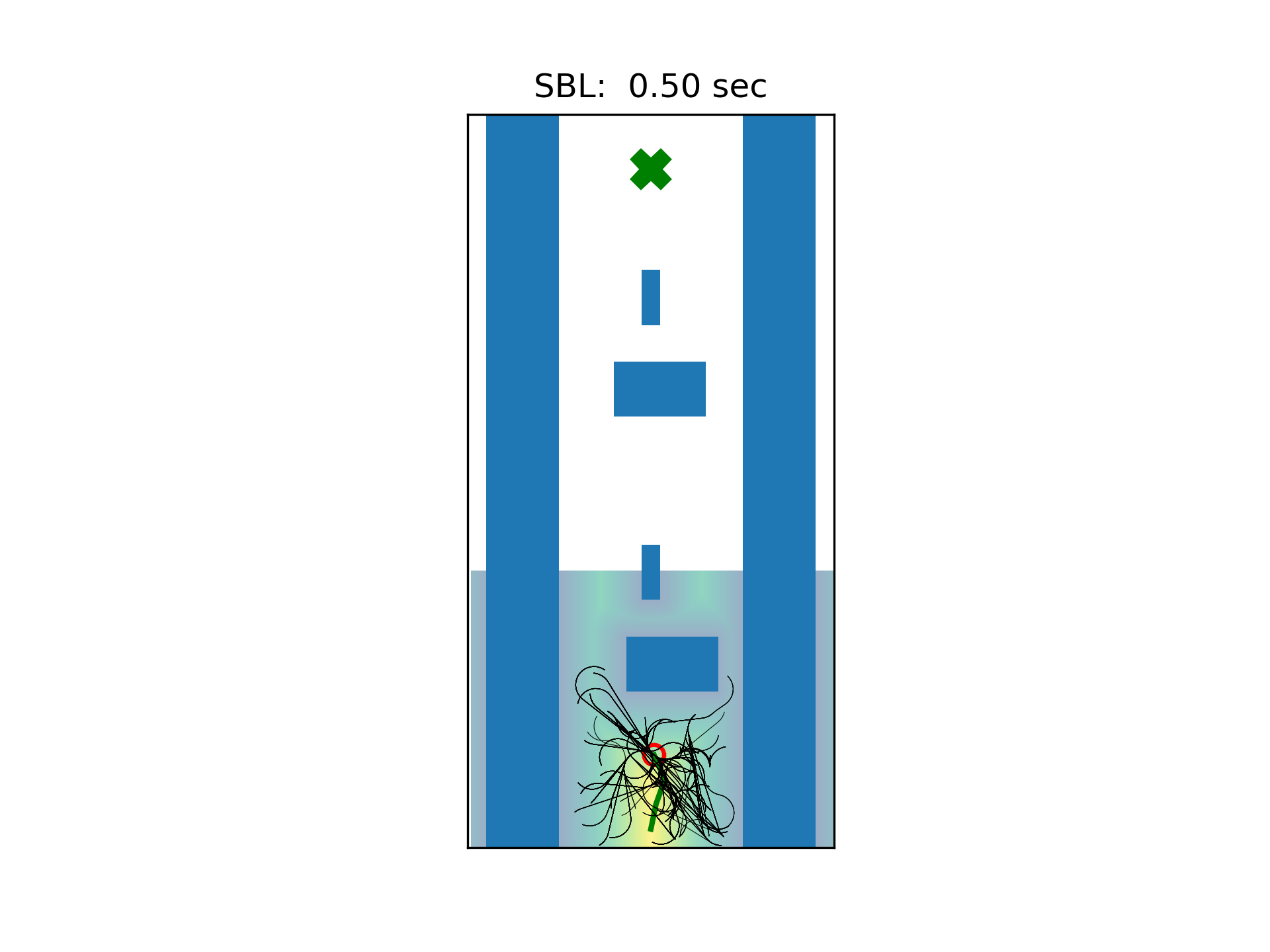}
  \hspace{-0.1cm}
  \includegraphics[trim=177 35 165 25,clip,width=0.08\linewidth]{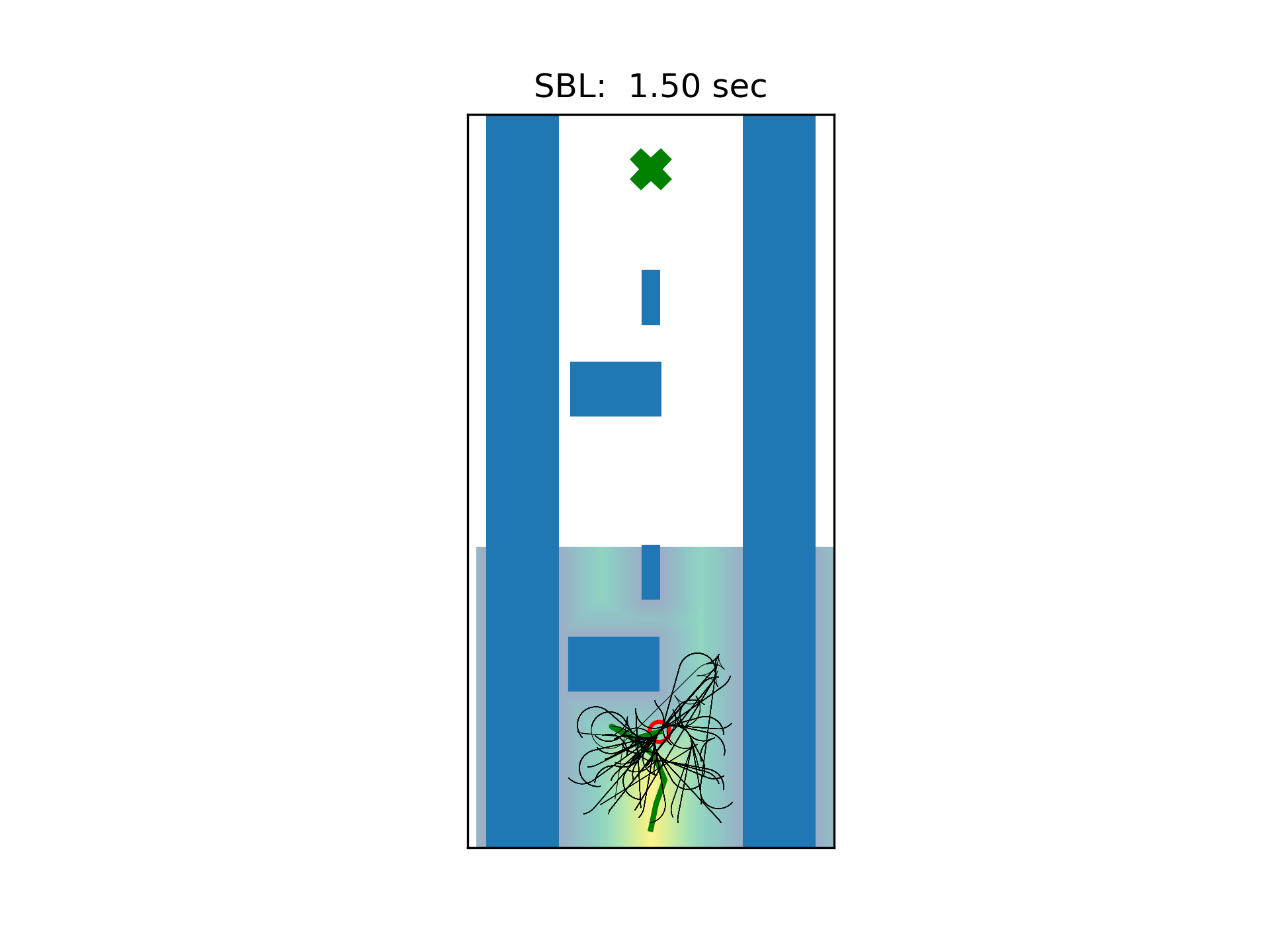}
  \hspace{-0.1cm}
  \includegraphics[trim=177 35 165 25,clip,width=0.08\linewidth]{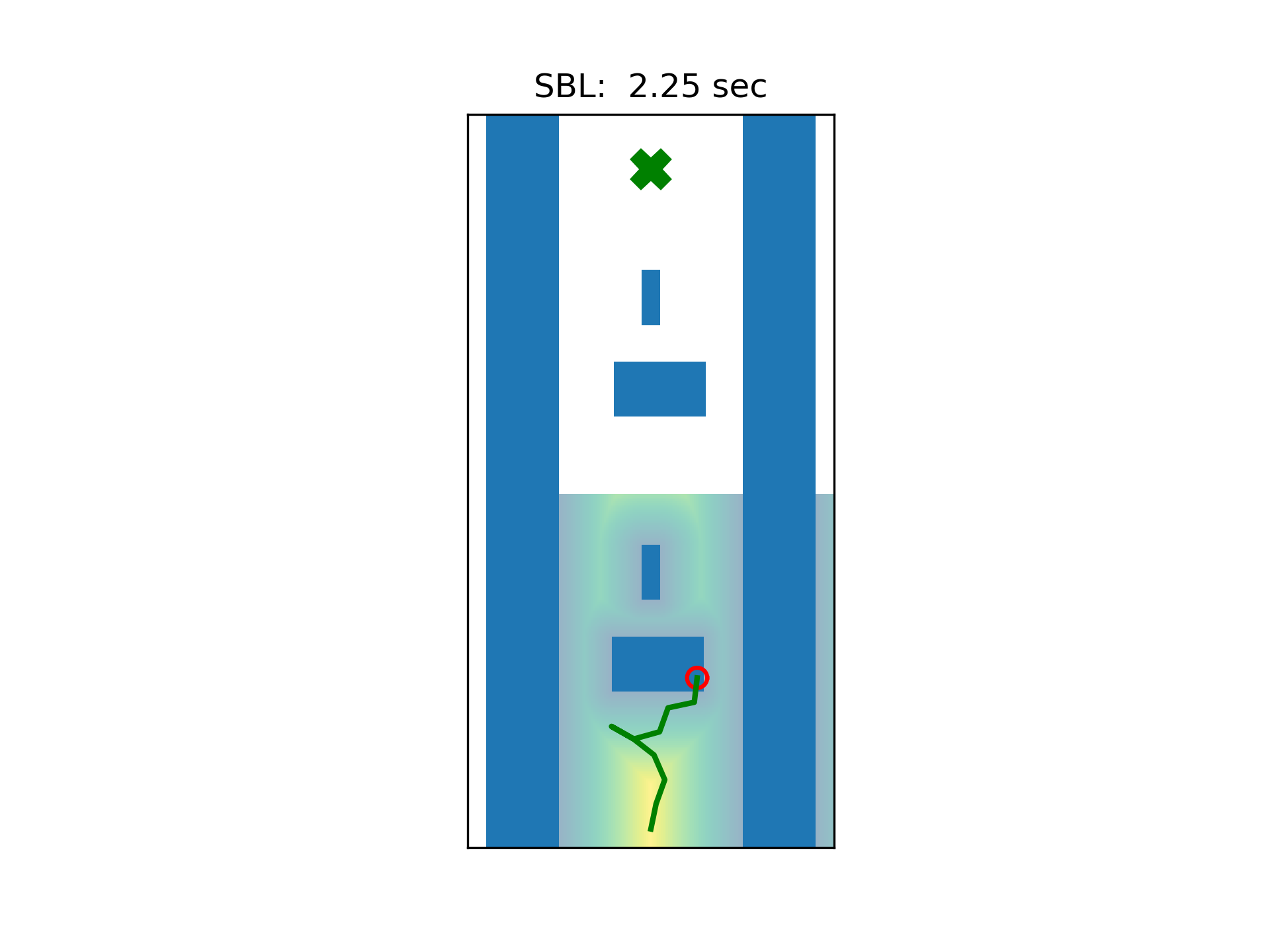}
  \captionof{figure}{\small Same Patrol benchmark example (left to right) where \ours (ours) (left) successfully reaches the goal while \blopt (middle) and \blsam (right) terminate in collision.}
  \label{fig:patrol}
  \vspace{-6mm}
\end{figure*}

For searching over the optimized factor graph and choosing the best path and action to execute, we employ a shortest path search where we pick the action or path leading to the leaf robot state on the factor graph with the lowest cost. The cost of any leaf robot state in the factor graph is computed as a summation of cost of all factors on the path leading up to the leaf robot state normalized by the depth of the leaf robot state to account for branches with different depths.

%%%%%%%%%%%%%%%%%%%%%%%%%%%%%%%%%%%%%%%%%%%%%%%%%%%%%%%%%%%%%%%%%
\vspace{-1mm}
\section{Evaluation}\label{sec:evaluations}
\vspace{-1mm}

We benchmark our approach \textbf{\ours} that combines sampling and optimization against a sampling only baseline (\textbf{\sbl}) and an optimization only baseline (\textbf{\obl}), in different highly dynamic environments in the domains of robot navigation and manipulation.

\textbf{Baselines} 
For the optimization baseline \obl, we setup GPMP2's~\cite{Mukadam-IJRR-18} single chain-like factor graph to operate in a receding horizon fashion that resembles MPC~\cite{Mukadam-ICRA-17}, but in state space. This serves as a stand-in for state-of-the-art in fast optimization based planning that can only reason over one hypothesis solution at a time. \ours reduces to \obl without the sampling enabled exploration and the resulting tree-like factor graph that holds multiple potential hypothesis to be optimized. On the other hand, the sampling baseline \sbl, is based on dynamic variants of RRT*~\cite{adiyatov2017novel} modified to operate in a receding horizon style to work in large environments (i.e. where start and goal can be far apart) and serves as a stand-in for state-of-the-art in sampling based methods. Any iteration of \sbl starts with the previous tree after pruning invalid and unreachable nodes and edges. The new samples are added while rewiring the tree towards the goal and terminates after hitting the node budget return the branch getting closest to the goal. \ours reduces to \sbl without running the optimization and instead checking node and edge validity while rewiring the tree. The same cost functions and factors are used in \obl and \sbl as used in \ours and all algorithms are executed under the same environmental conditions and robot constraints. \ours and \obl plan in the space of position and velocity of the robot, while \sbl only plans in space of robot position to which average velocity is applied for execution.

\textbf{Environments} In our benchmark environments, \textbf{2D Static}, \textbf{2D Forest}, \textbf{Patrol}, \textbf{Pedestrian}, and \textbf{3D Forest}, we assume that the robot is only aware of the locations of obstacles in a local neighborhood (that corresponds to real world LIDAR ranges) and not their velocities. We also incorporate stochastic actions and noisy state measurements for the robots (with Gaussian noise). In all our experiments, we individually tuned all the hyperparameters in each algorithm to get their best performance.

\textbf{Metrics} We record the following metrics that are averaged over a set of trials, run with unique random seeds shared by all algorithms such that they encounter the exact same set of planning problems (start, goal, and environment): (i) rate of success of reaching goal without getting into collisions or timing out (\textbf{Success}), (ii) total execution i.e. wall-clock time taken by the robot to reach the goal from the start (\textbf{Execution time}) averaged over successful runs, (iii) average computation time per iteration taken by the algorithm to plan (\textbf{Compute time}) measured starting from the input from the perception module (e.g. signed distance field of the environment) to the output plan from the algorithm, and (iv) total distance traveled by the robot normalized by the starting euclidean distance to goal (\textbf{Norm. Dist}) averaged over successful runs.

%---------------------------------------------------------------
\subsection{2D Static: Static Benchmark for Navigation}\label{sec:2dstatic}
\vspace{-1mm}

We begin with a benchmark in a static setting to establish the starting performance of our baselines and show that they excel here, as discussed in Section~\ref{sec:intro}, as well as demonstrate that our method performs competitively in this setting. We consider the task of planar point goal (x, y, yaw) navigation in a large environment with even distribution of uniformly spaced static square shaped obstacles on a grid. In this environment, we consider a differential drive robot (non-holonomic constraint) that is handled in \ours and \obl with a unary factor on any state while for \sbl we use Reeds-Shepp motion curves ~\cite{reeds1990optimal} when extending and connecting states on the tree. We also enforce velocity limits on the robot for all algorithms (as factors for the former two). 30 random trials are generated where the start and goal are at least a fixed minimum distance apart.

The benchmark results are summarized in \tref{table:env} (row 1). We observe that all methods are successfully able to reach the goal in every run and the difference in the other metrics reveal the underlying key algorithmic differences. Sampling enabled exploration allows \ours and \sbl to reach the goal more quickly (lower execution time). The optimization property leads to lower compute times for \ours and \obl along with smoother trajectories and lower normalized distances compared to \sbl. \ours naturally incurs a slightly higher compute time since it is optimizing a more complex tree shaped factor graph with more connection compared to a single chain like factor graph of \obl.

\begin{figure}[!t]
\centering
\includegraphics[trim=200 250 900 500,clip, width=0.85\linewidth]{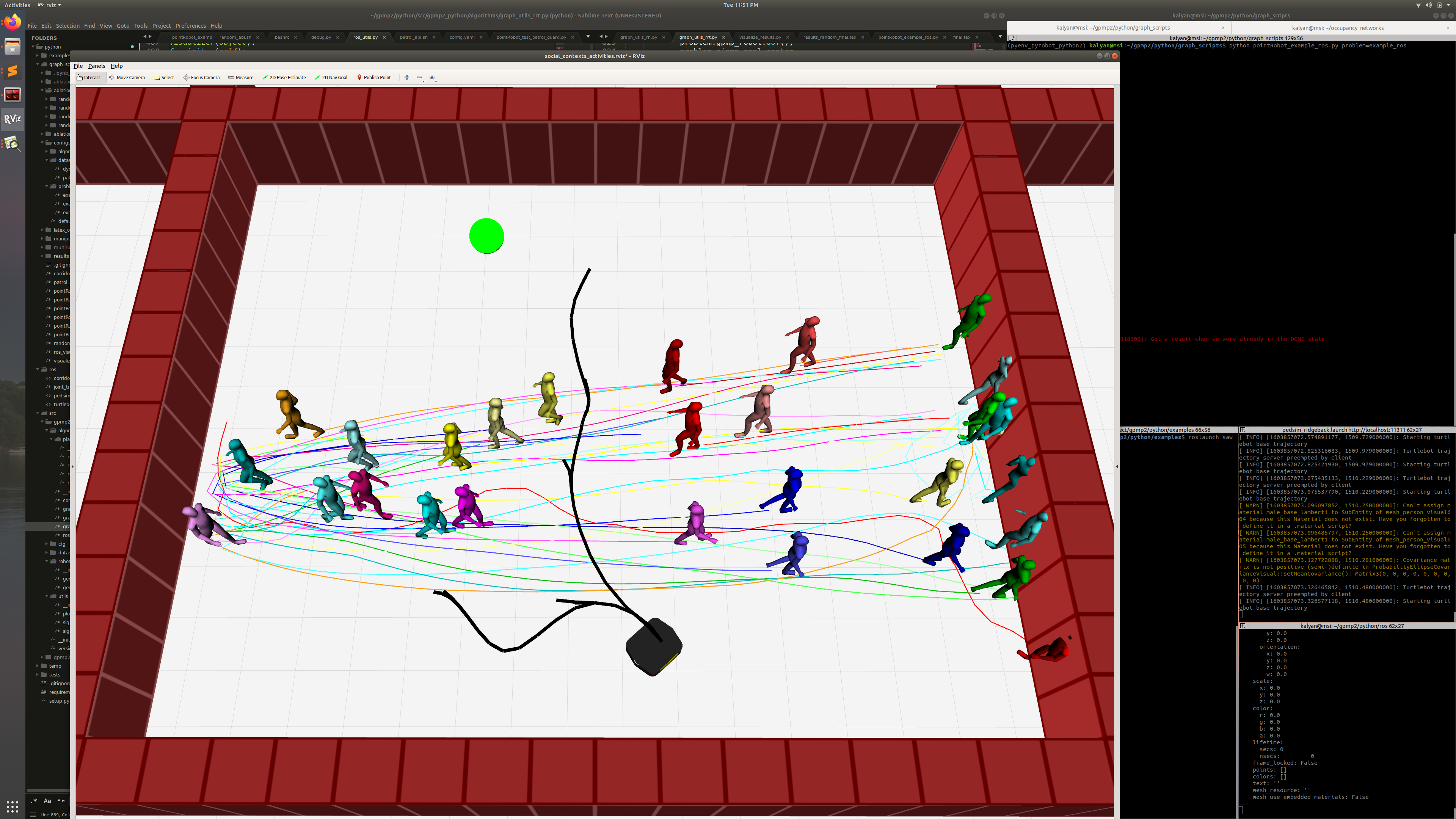}
\caption{\small Pedestrian benchmark example in Gazebo with \ours (ours).}
\label{fig:gazebo}
\vspace{-2mm}
\end{figure}

\begin{figure}[!t]
\centering
\begin{tikzpicture}
\draw (0, 0) node[inner sep=0] {
\includegraphics[trim=25cm 20cm 60cm 20cm,clip,width=0.48\linewidth]{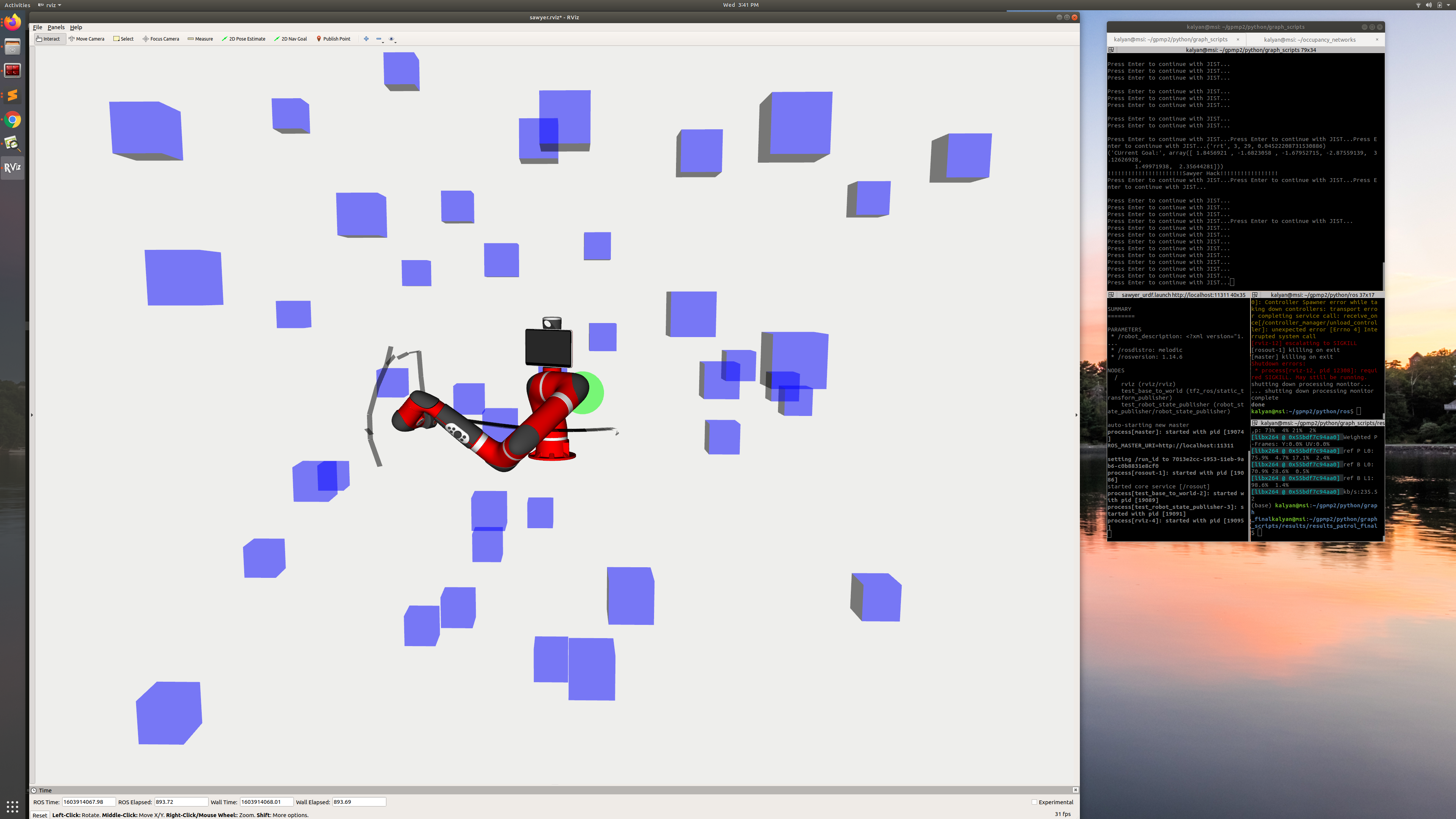}
\includegraphics[trim=25cm 20cm 60cm 20cm,clip,width=0.48\linewidth]{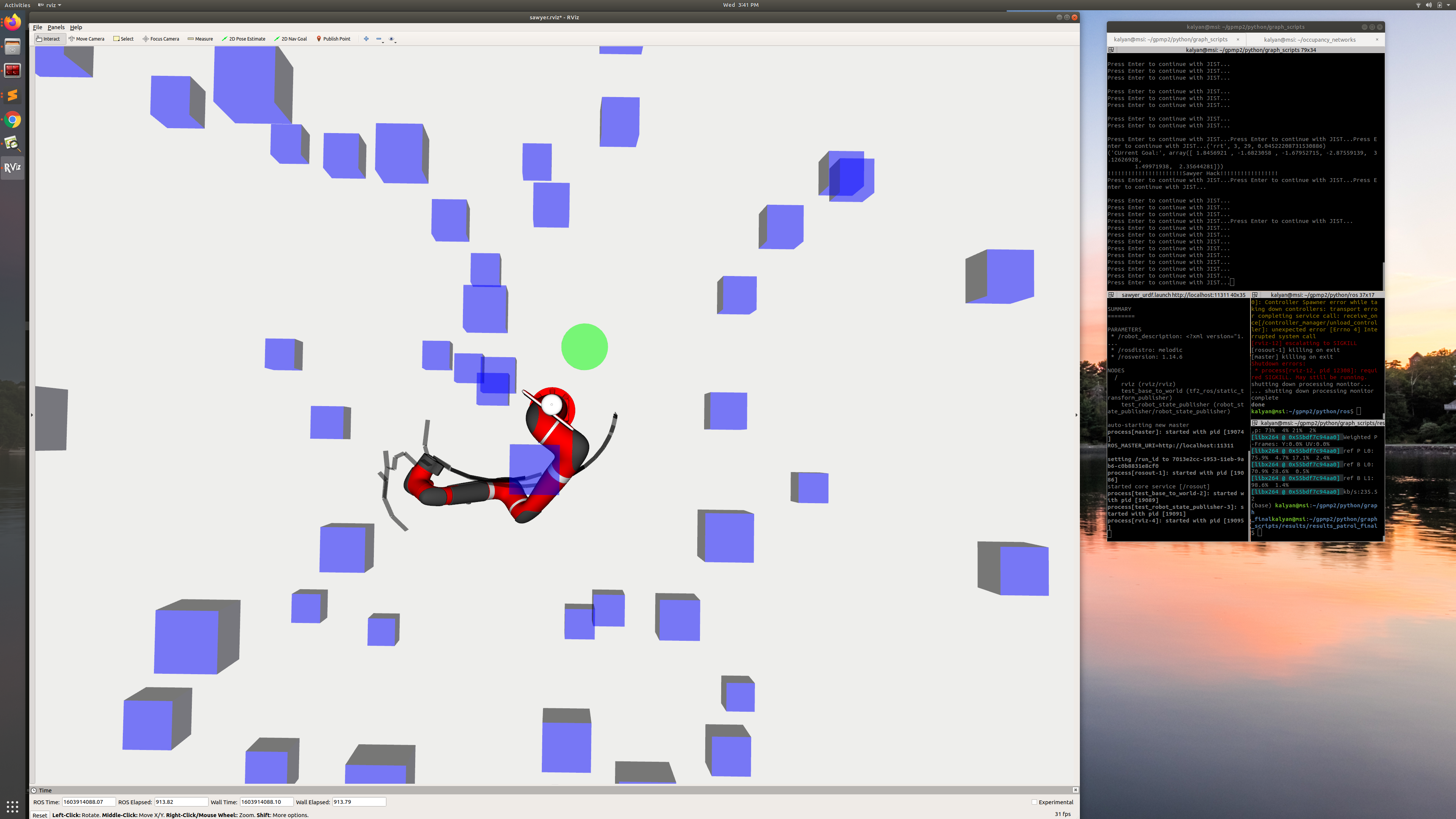}
};
\draw (-0.8, 1.3) node {\small front view};
\draw (0.7, 1.27) node {\small top view};
\end{tikzpicture}
\caption{\small 3D Forest benchmark example with \ours (ours).}
\label{fig:manipulation}
\vspace{-1mm}
\end{figure}

\begin{table*}[!t]
\centering
\caption{\small Benchmark results on all environments.}
\vspace{-1mm}
\label{table:env}
% \resizebox{1.5\columnwidth}{!}{
% \begin{tabu}{|c|[1pt]c|c|c|[1pt]c|c|c|[1pt]c|c|c|[1pt]c|c|c|}%
\begin{tabu}{|l|c|c|c|c|c|c|c|c|c|c|c|c|}%

\hline%
&\multicolumn{3}{c|}{\textbf{Success}}&\multicolumn{3}{c|}{\textbf{Execution time (s)}}&\multicolumn{3}{c|}{\textbf{Compute time (s)}}&\multicolumn{3}{c|}{\textbf{Norm. Dist}}\\%
% \hline%
&\textbf{\ours}&\textbf{\obl}&\textbf{\sbl}&\textbf{\ours}&\textbf{\obl}&\textbf{\sbl}&\textbf{\ours}&\textbf{\obl}&\textbf{\sbl}&\textbf{\ours}&\textbf{\obl}&\textbf{\sbl}\\%

\hline%
\hline%
2D Static&1.00&1.00&1.00&28.08&51.53&34.67&0.016&0.011&0.252&1.38&1.41&1.65\\%

% \hline%
2D Forest&0.43&0.20&0.03&47.75&75.50&17.75&0.016&0.011&0.325&1.95&1.98&1.88\\%

% \hline%
Patrol&0.60&0.27&0.07&18.32&27.03&7.75&0.166&0.385&0.451&1.26&1.18&2.07\\%

% \hline%
Pedestrian&0.70&0.45&0.00&288.04&170.53&-&0.039&0.035&-&4.94&2.15&-\\%

% \hline%
3D Forest&0.71&0.60&0.36&18.64&18.69&11.83&0.140&0.256&0.403&5.37&3.29&6.51\\%
\hline%
\end{tabu}
% }
\end{table*}

\begin{table*}[!t]
\centering
\vspace{-2mm}
\caption{\small Ablations on 2D Forest environment.}
\vspace{-2mm}
\label{table:abl}
% \resizebox{\columnwidth}{!}{%
% \begin{tabu}{|c|[1pt]c|c|c|[1pt]c|c|c|[1pt]c|c|c|[1pt]c|c|c|}%
\begin{tabu}{|c|c|c|c|c|c|c|c|c|c|c|c|c|}%
% \hline%

\multicolumn{13}{c}{\textbf{Environment obstacle number ablation}}\\%
\hline%
&\multicolumn{3}{c|}{\textbf{Success}}&\multicolumn{3}{c|}{\textbf{Execution time (s)}}&\multicolumn{3}{c|}{\textbf{Compute time (s)}}&\multicolumn{3}{c|}{\textbf{Norm. Dist}}\\%
% \hline%
&\textbf{\ours}&\textbf{\obl}&\textbf{\sbl}&\textbf{\ours}&\textbf{\obl}&\textbf{\sbl}&\textbf{\ours}&\textbf{\obl}&\textbf{\sbl}&\textbf{\ours}&\textbf{\obl}&\textbf{\sbl}\\%
\hline%
\hline%
20&0.97&0.90&0.37&21.45&43.94&18.98&0.014&0.011&0.325&1.11&1.25&1.54\\%
% \hline%
40&0.80&0.67&0.13&28.27&59.38&19.12&0.015&0.011&0.304&1.32&1.58&1.78\\%
% \hline%
60&0.57&0.63&0.13&34.72&66.82&19.44&0.016&0.011&0.326&1.54&1.75&1.77\\%
% \hline%
80&0.43&0.20&0.03&47.75&75.50&17.75&0.015&0.011&0.320&1.95&1.98&1.88\\%
% \hline%
100&0.40&0.07&0.03&47.58&92.50&23.75&0.015&0.011&0.371&1.85&2.48&1.77\\
% \tabucline[1pt]{-}%
\hline%

\multicolumn{13}{c}{\textbf{Environment obstacle top speed ablation (m/s)}}\\%
\hline%
&\multicolumn{3}{c|}{\textbf{Success}}&\multicolumn{3}{c|}{\textbf{Execution time (s)}}&\multicolumn{3}{c|}{\textbf{Compute time (s)}}&\multicolumn{3}{c|}{\textbf{Norm. Dist}}\\%
% \hline%
&\textbf{\ours}&\textbf{\obl}&\textbf{\sbl}&\textbf{\ours}&\textbf{\obl}&\textbf{\sbl}&\textbf{\ours}&\textbf{\obl}&\textbf{\sbl}&\textbf{\ours}&\textbf{\obl}&\textbf{\sbl}\\%
\hline%
\hline%
0.5&0.63&0.53&0.30&78.61&135.83&24.69&0.015&0.011&0.388&2.46&2.60&2.17\\%
% \hline%
1.0&0.53&0.40&0.10&51.78&87.56&19.83&0.017&0.011&0.305&1.92&2.03&1.76\\%
% \hline%
1.5&0.43&0.20&0.03&47.75&75.50&17.75&0.016&0.011&0.325&1.95&1.98&1.88\\%
% \hline%
2.0&0.30&0.10&0.00&45.22&63.67&-&0.016&0.011&-&1.86&1.80&-\\%
% \hline%
2.5&0.40&0.17&0.03&40.29&56.90&20.50&0.016&0.011&0.329&1.83&1.74&1.74\\
% \tabucline[1pt]{-}%
\hline%

\multicolumn{13}{c}{\textbf{Execution noise ablation (m)}}\\%
\hline%
&\multicolumn{3}{c|}{\textbf{Success}}&\multicolumn{3}{c|}{\textbf{Execution time (s)}}&\multicolumn{3}{c|}{\textbf{Compute time (s)}}&\multicolumn{3}{c|}{\textbf{Norm. Dist}}\\%
% \hline%
&\textbf{\ours}&\textbf{\obl}&\textbf{\sbl}&\textbf{\ours}&\textbf{\obl}&\textbf{\sbl}&\textbf{\ours}&\textbf{\obl}&\textbf{\sbl}&\textbf{\ours}&\textbf{\obl}&\textbf{\sbl}\\%
\hline%
\hline%
0.0&0.43&0.27&0.00&43.06&66.75&-&0.015&0.010&-&1.67&1.77&-\\%
% \hline%
0.05&0.33&0.23&0.10&40.77&70.46&16.67&0.015&0.010&0.268&1.61&1.95&1.72\\%
% \hline%
0.1&0.40&0.23&0.13&43.38&82.68&22.44&0.015&0.010&0.329&1.72&1.95&1.87\\%
% \hline%
0.15&0.60&0.17&0.10&51.15&166.55&19.42&0.015&0.010&0.260&1.99&2.15&1.81\\%
% \hline%
0.2&0.50&0.17&0.07&46.60&137.50&21.12&0.015&0.011&0.263&1.82&2.46&2.08\\
% \tabucline[1pt]{-}%
\hline%

\multicolumn{13}{c}{\textbf{Node budget ablation}}\\%
\hline%
&\multicolumn{3}{c|}{\textbf{Success}}&\multicolumn{3}{c|}{\textbf{Execution time (s)}}&\multicolumn{3}{c|}{\textbf{Compute time (s)}}&\multicolumn{3}{c|}{\textbf{Norm. Dist}}\\%
% \hline%
&\textbf{\ours}&\textbf{\obl}&\textbf{\sbl}&\textbf{\ours}&\textbf{\obl}&\textbf{\sbl}&\textbf{\ours}&\textbf{\obl}&\textbf{\sbl}&\textbf{\ours}&\textbf{\obl}&\textbf{\sbl}\\%
\hline%
\hline%
40&0.43&0.30&0.07&45.87&77.83&20.12&0.010&0.007&0.188&1.69&1.97&1.53\\%
% \hline%
60&0.43&0.20&0.03&47.75&75.50&17.75&0.016&0.011&0.324&1.95&1.98&1.88\\%
% \hline%
80&0.37&0.27&0.03&39.93&61.53&21.00&0.020&0.013&0.429&1.75&1.68&2.14\\%
% \hline%
100&0.53&0.27&0.00&41.42&59.00&-&0.027&0.017&-&1.83&1.70&-\\%
% \hline%
120&0.43&0.23&0.07&33.79&74.14&19.20&0.034&0.021&0.496&1.74&2.10&1.89\\%
\hline%

\end{tabu}

% }
\vspace{-5mm}
\end{table*}

%---------------------------------------------------------------
\subsection{2D Forest: Random Forest Benchmark for Navigation}\label{sec:random_env}
\vspace{-1mm}

The primary benchmark task we consider is planar point goal (x, y, yaw) navigation in a highly dynamic 2D forest with large number of randomly moving square shaped obstacles that can accelerate in random directions as shown in \fref{fig:random}. We use the same differential drive robot and the same costs and factors for the algorithms from the 2D Static benchmark. The robot's limited visibility is also shown by the shaded square in \fref{fig:random} that represents the instantaneous signed distance field (SDF) used for collision checking.

The benchmark results are summarized in \tref{table:env} (row 2). Under the increased difficulty of the problem compared to the static setting, we observe drop in performance for all algorithms. But, \ours is significantly more successful at reaching the goal compared to both the baselines and reaching the goal faster compared to \obl with nearly the same amount of computation time differences as in the 2D Static benchmark. This shows that the \ours is able to leverage the exploration enabled by sampling but isn't burdened by a static graph that is more expensive to update. Instead it can exploit optimization to quickly reason over the multiple potential plans. On the other hand, \sbl retains a static tree and only removes nodes or edges in collision, but cannot move them a safe distance away from obstacles as done during optimization. \obl cannot explore and is only able to optimize and reason over only one potential plan.

To study how the performance gap between the three algorithms varies as the problem difficulty scales, we also carry out ablation studies in this environment and are presented in \tref{table:abl}. First, we vary the number of dynamic obstacles and their speeds to scale the difficulty. We observe that starting at the lowest difficulty that is closest to the static setting \sbl already drops in success. As the problem becomes more difficult the success rate of the baselines falls close to zero, while \ours only gradually falls to $0.4$. Increasing difficult also increases the execution time and normalized distance, but all algorithms maintain similar differences in their compute times. The outliers are due to the small number of successful outcomes in those settings. 
Then, we vary the variance of noise in measuring the current state of the robot and the noise in executing actions. All algorithms are able to maintain their success rate given the benefits of replanning, but incur slightly higher execution times and normalized distance under high noise. Finally, we vary the available node budget that corresponds to the cap on the maximum number of nodes in the tree for \ours and \blsam, and the length of horizon for \blopt. We observe that the computation times increases with increased node budget with minor improvements in success rate. Later in Section~\ref{sec:dis} we discuss alternate ways of sampling that can be explored to improve performance.

%---------------------------------------------------------------
\vspace{-1mm}
\subsection{Patrol: Dynamic Narrow Passage Benchmark}
\vspace{-1mm}

Next, we benchmark all three algorithms in the setting of robot navigation through a narrow passages guarded by dynamic obstacles as shown in \fref{fig:patrol} and is a more complex version of the patrol guard environment from~\cite{kolur2019online}. Here the passage consists of two rectangular obstacles moving back and forth along a horizontal line with random velocity at two different locations in the passage. For this environment, we use the same robot agent, algorithm specific considerations, and noise models as we did in the 2D Forest benchmarks. In these experiments we randomly sample a start and goal at both ends of the passage respectively. Results averaged over 30 trials are summarized in \tref{table:env} (row 3). Our findings are in the vein of the previous benchmarks where we observe \ours is able to outperform in success rate over the baseline while incurring a small added computation time over \obl which is still overall much lower compared to \sbl.

%---------------------------------------------------------------
\vspace{-0.5mm}
\subsection{Pedestrian: Benchmark in Gazebo Simulator}
\vspace{-1mm}

We also perform a benchmark in a more practical navigation setting where the robot has to reach a goal while avoiding pedestrians as shown in \fref{fig:gazebo}. In this setting, we make the experiments realistic and real-time by deploying the entire system in the Gazebo~\cite{koenig2004design} physics simulator. We simulate the robot's perception to build SDFs from LIDAR scans and simulate the robot's control for execution using PID velocity controllers. We use the ClearPath Ridgeback mobile robot that has an omni-directional drive and do not use the holonomic constraint, but still enforce velocity limits. We also simulate realistic pedestrian motion in crowds using Pedsim~\cite{gloor2016pedsim}. The results from this benchmark are averaged over 100 trials and presented in \tref{table:env} (row 4) and shows \ours similarly outperforms both baselines.

%---------------------------------------------------------------
\subsection{3D Forest: Random Forest Benchmark for Manipulation}
\vspace{-1mm}

Lastly, we benchmark a manipulation task as shown in \fref{fig:manipulation}. Here, we consider online planning for a seven degree of freedom Sawyer robot arm in a dynamic environment with cubes moving with random velocities. Robot joint limit and self-collision avoidance is enforced for all algorithms (as factors for \ours and \blopt). The robot has full visibility of its workspace since the manipulator base remains fixed. The goal for the robot is specified as a target end-effector location. For the experiments we randomly sample a start joint configuration for the robot arm and a reachable task space goal location. The normalized distance metric is computed using the task space distance traveled by the end-effector. We present the averaged results over 100 trials in \tref{table:env} (row 5) and find results consistent with the other benchmarks.

%%%%%%%%%%%%%%%%%%%%%%%%%%%%%%%%%%%%%%%%%%%%%%%%%%%%%%%%%%%%%%%%%
\vspace{-1mm}
\section{Discussion}\label{sec:dis}
\vspace{-1mm}

In this work, we focused on the problem of motion planning in large, dense, highly dynamic environments and identified the essence of the problem as the ability to efficiently grow, track, and switch between multiple possible candidate solutions. We validated this with our approach \ours that combines the exploration and exploitation strengths of sampling and optimization respectively. In this approach we do not check the samples or the edge connections made during sampling for collision and instead let the optimizer try to move the samples and the edges out of collision. In doing so we gain practical efficiency but sacrifice the theoretical properties like probabilistic completeness offered by sampling methods.

We currently precompute the SDFs and cache them for the 3D environments. While SDFs can be computed online for 2D environments, in 3D settings, solutions to building SDFs incrementally online on GPUs~\cite{oleynikova2017voxblox} can be utilized to support the perception input for our approach and maintain real time requirements for the overall system.

While in this specific implementation we used random sampling and nonlinear least square optimization, our framework is flexible enough to support alternate sampling and optimization strategies with the underlying factor graph acting as the core data structure tying them together. To mitigate local minima or bias sampling toward other desired outcomes it is possible to use goal directed sampling~\cite{gammell2014informed}, learn heuristics~\cite{choudhury2018data} or learn to sample~\cite{ichter2018learning}. Similarly, optimization can be adapted to support learned cost functions and weights~\cite{bhardwaj2020differentiable}.

%%%%%%%%%%%%%%%%%%%%%%%%%%%%%%%%%%%%%%%%%%%%%%%%%%%%%%%%%%%%%%%%%
\vspace{-2mm}
\bibliographystyle{IEEEtran}
\bibliography{IEEEabrv,ref}

%%%%%%%%%%%%%%%%%%%%%%%%%%%%%%%%%%%%%%%%%%%%%%%%%%%%%%%%%%%%%%%%%
% {\ifNOTARXIV\clearpage\fi}
{\ifAPP
\section*{Appendix}

\subsection{Factor Descriptions}

In this section we provide details of all the factors used in the \ours factor graph in our current experiments.

\textbf{GP prior factor} ensure smoothness in the resulting trajectory and encodes the optimality criteria in the objective function~\cite{Mukadam-IJRR-18}. It is derived from a linear time varying system~\cite{barfoot2014batch} meant to represent the motion model of the robot.
\begin{gather*}
\bm{f}^{gp} (\bm\theta^i_t, \bm\theta^j_{t+1}) = \exp \Big\{ \hspace{-1mm}-\frac{1}{2} \| \mathbf{\Phi}_{t,t+1} \bm\theta^i_t - \bm\theta^j_{t+1} \|^2_{\mathbf{Q}_{t,t+1}} \Big\} \\
\small{
\mathbf{\Phi}_{t,t+1} = \begin{bmatrix}
\mathbf{I} & \Delta t\mathbf{I} \\ 
\mathbf{0} & \mathbf{I}
\end{bmatrix}, \quad
\mathbf{Q}_{t,t+1} = \begin{bmatrix}
\frac{1}{3} \Delta t^3 \mathbf{Q}_c &
\frac{1}{2} \Delta t^2 \mathbf{Q}_c \\ 
\frac{1}{2} \Delta t^2 \mathbf{Q}_c &
\Delta t \mathbf{Q}_c
\end{bmatrix}
}
\end{gather*}
where $\mathbf{\Phi}_{t,t+1} $ is the state transition matrix during time interval $\Delta t$ between $t$ and $t+1$, with variance $\mathbf{Q}_{t,t+1} $ and hyperparameter $\mathbf{Q}_c$.

\textbf{Obstacle avoidance factor} on any state is used to keep that state collision free from obstacles in the environment.
\begin{gather*}
\bm{f}^{obs} (\bm\theta^i_t) = \exp \Big\{ \hspace{-1mm}-\frac{1}{2} \| \mathbf c(\mathbf{d}(\bm\phi( \bm\theta^i_t ))) \|^2_{\bm{\Sigma}^{obs}} \Big\}
\end{gather*}
where $\bm{\Sigma}^{obs} = \sigma_{obs}^2 \mathbf{I}$ and $\bm\phi$ maps the robot configuration to a set of representative points on the robot body in task space that are queried in the SDF $\mathbf{d}$ to get their signed distances. Hinge loss cost function $\mathbf{c}$ with safety distance $\epsilon$ then gives the obstacle collision cost given the signed distances~\cite{Mukadam-IJRR-18}.

\textbf{Binary obstacle avoidance factor} between two consecutive states uses fast GP interpolation~\cite{barfoot2014batch} $\mathbf{p}$ to find the state at time $\tau$ in the interval $(t,t+1)$ and the same hinge loss cost for collision avoidance. A set of these together in between states ensures that the edge is also evaluated for collision~\cite{Mukadam-IJRR-18}.
\begin{gather*}
\small{
\bm{f}^{intp} (\bm\theta^i_t, \bm\theta^j_{t+1}; \tau) = \exp \Big\{ \hspace{-1mm}-\frac{1}{2} \| \mathbf c(\mathbf{d}(\bm\phi( \mathbf{p}(\bm\theta^i_t, \bm\theta^j_{t+1}, \tau )))) \|^2_{\bm{\Sigma}^{obs}} \Big\}
}
\end{gather*}

\textbf{Start or current state factor} is used to ground the trajectory at the start or current state and also represents the measurement $\bm\theta_{curr}$ of the current state with variance $\bm{\Sigma}^{curr} = \sigma_{curr}^2 \mathbf{I}$.
\begin{gather*}
\bm{f}^{curr} (\bm\theta^i_t) = \exp \Big\{ \hspace{-1mm}-\frac{1}{2} \| \bm\theta^i_t - \bm\theta_{curr} \|^2_{\bm{\Sigma}^{curr}} \Big\}
\end{gather*}

\textbf{Goal factor} is used on every state except the current state to drive all path towards the goal. The strength of the factor is based on the proximity of the current state to the goal~\cite{Mukadam-ICRA-17}.
\begin{gather*}
\bm{f}^{goal} (\bm\theta^i_t) = \exp \Big\{ \hspace{-1mm}-\frac{1}{2} \| \bm\theta^i_t - \bm\theta_{goal} \|^2_{\bm{\Sigma}^{goal}} \Big\} \\
\bm{\Sigma}^{goal} = \sigma_{goal}^2  \frac{\|\bm\theta_{curr} - \bm\theta_{goal}\|^2}{\| \bm\theta_{start} -  \bm\theta_{goal}\|^2} \mathbf{I}
\end{gather*}

\textbf{Joint and velocity limit factor} are enforced with a similar hinge loss cost function like obstacle avoidance where a penalty is incurred if the state gets $\epsilon$ close or over the limit in any dimension of the state.
\begin{gather*}
\bm{f}^{lim} (\bm\theta^i_t) = \exp \Big\{ \hspace{-1mm}-\frac{1}{2} \| \mathbf{c} (\bm\theta^i_t) \|^2_{\bm{\Sigma}^{lim}} \Big\} \\
c =
\begin{cases} 
\hfill ll + \epsilon - \theta^i_t \hfill & \text{if} \ \theta^i_t < ll + \epsilon  \\
\hfill \theta^i_t - ul - \epsilon \hfill & \text{if} \ ul + \epsilon < \theta^i_t \\
\hfill 0 \hfill & \text{if} \ ll + \epsilon \leqslant \theta^i_t \leqslant ul + \epsilon \\
\end{cases}
\end{gather*}
where $c$ evaluates the cost for each scalar dimension $\theta^i_t$ of state vector $\bm\theta^i_t$ and $\mathbf{c}$ stacks all $c$ together. $ul$ and $ll$ are the upper and lower limits respectively and $\bm{\Sigma}^{lim} = \sigma_{lim}^2 \mathbf{I}$.

\textbf{Non-holonomic constraint factor} is used for planar robots with state $\bm\theta^i_t = [x^i_t, y^i_t, \psi^i_t, \dot{x}^i_t, \dot{y}^i_t, \dot{\psi}^i_t]^\top$ with a differential drive system to restrict implausible sideways motion of the robot.
\begin{gather*}
\bm{f}^{nh} (\bm\theta^i_t) = \exp \Big\{ \hspace{-1mm}-\frac{1}{2} \| \dot{y}^i_t \cos(\psi^i_t) - \dot{x}^i_t \sin(\psi^i_t) \|^2_{\sigma_{nh}^2} \Big\}
\end{gather*}

\textbf{Self collision avoidance factor} is useful for manipulators with high degree of freedom to avoid self collision. We use a similar hinge loss cost function like the obstacle avoidance except here the distance function $\mathbf{d}$ calculates the distances of all representative points on the robot body with respect to each other. The pairs of points to be checked are pre-specified and optimized to not check pairs of points for instance that are on the same link of the robot.
\begin{gather*}
\bm{f}^{self} (\bm\theta^i_t) = \exp \Big\{ \hspace{-1mm}-\frac{1}{2} \| \mathbf c(\mathbf{d}(\bm\phi( \bm\theta^i_t ))) \|^2_{\bm{\Sigma}^{self}} \Big\}
\end{gather*}
where $\bm{\Sigma}^{self} = \sigma_{self}^2 \mathbf{I}$.

\subsection{Experimental Details}
In this section we provide additional details on the experiments.

\subsubsection{Environments and robots}

In all navigation experiments the top allowed speed for the robots is 3m/s for translation and 0.6rad/s for rotation.
\textbf{2D Static} and \textbf{2D Forest} environments are 90m $\times$ 120m in size. All obstacles are square shaped with side length 6m and move in the environment with random acceleration limited between -0.6m/s\textsuperscript{2} and 0.6m/s\textsuperscript{2} in random directions (zero velocity and acceleration in static case). 2D Static environment comprised of 48 obstacles uniformly spaced at 15m on a grid and the 2D Forest consisted of 80 randomly distributed dynamic obstacles. The robot is circular with a radius of 1.5m. We considered a Gaussian noise for both the robot dynamics model and robot localization model with a mean of [0.03m, 0.03m, 0.03 radians]. 
\textbf{Patrol} environment is 10m $\times$ 40m in size with two moving and two stationary rectangular obstacles. Robot is the same as in 2D Forest but with a smaller radius of 0.5m.
\textbf{Pedestrian} environment is 14m $\times$ 15m in size. The pedestrians have a top speed of 2m/s and follow a swarm-like crowd motion, simulated using~\cite{gloor2016pedsim}. At any given point, the number of pedestrians on the map range from 15 to 40 in our experiments and is not fixed as the simulator lets pedestrians enter and leave in a continuous stream.
\textbf{3D Forest} environment consists of 50 randomly moving cubical obstacles of size length 0.15m with top speed of 0.5m/s. The obstacles are prevented from colliding with the base link of the manipulator as it is fixed in the environment. The top allowed speed for all joints of the manipulator is 1rad/s

\subsubsection{Algorithms}

Hyperparameters of all factors and cost functions are tuned individually for each algorithm. Constraints are handled with factors for \ours and \blopt, while for \blsam we perform explicit collision checking, velocity limit check, robot dynamics and kinematic constraint validation while adding new states and edges to the tree. To produce dynamically efficient paths that are also at a safe distance away from the obstacles, in \blsam, we use the same cost function terms that we used to model the GP factor and obstacle avoidance factor. We also bias the search towards the goal for \blsam by simply augmenting the cost function with a goal heuristic (weighted euclidean distance to the goal) from any given state on the tree. 
Unlike the configuration position and velocity based state in \ours and \blopt, the state in \blsam only consists of configuration position. Thus, \blsam provides a path that we then post process with linear interpolation to fit a trajectory.
In navigation experiments, while growing the trees in \ours and \blsam, we limit the sampling to a local neighborhood around the robot. This encourages the algorithms to explore more in the local visible neighborhood of the robot which has a higher influence on its immediate future.
\fi}

%%%%%%%%%%%%%%%%%%%%%%%%%%%%%%%%%%%%%%%%%%%%%%%%%%%%%%%%%%%%%%%%%

\end{document}